\documentclass{article}

\usepackage{genai4health_2026}   

\usepackage[utf8]{inputenc}
\usepackage[T1]{fontenc}
\usepackage{hyperref}
\usepackage{url}
\usepackage{booktabs}
\usepackage{amsfonts}
\usepackage{amsmath}
\usepackage{nicefrac}
\usepackage{microtype}
\usepackage{xcolor}
\usepackage{graphicx}
\usepackage{multirow}
\usepackage{placeins}
\usepackage{float}
\usepackage{pifont}     
\usepackage{amssymb}    

\title{PRISM: Pathology Reliability In Scarce-label Medicine.\\[3pt]
{\Large A Tile-level Benchmark of Pathology Foundation Models}}

\author{%
  Alicem Koyun \\
  Westcliff University \\
  \texttt{alicemkoyun@gmail.com} \\
}

\begin{document}

\maketitle

\begin{abstract}
Pathology foundation models are typically benchmarked on classification
performance at full label availability and on in-distribution test sets, while
the properties relevant to clinical deployment (calibration, behavior at low
label budgets, and robustness under cross-institutional shift) are seldom
evaluated jointly. We introduce \textbf{PRISM}, a tile-level benchmark that
measures the reliability of pathology foundation models along four axes
simultaneously: model choice, label fraction, calibration, and
out-of-distribution transfer. PRISM evaluates eight foundation models (CLIP,
PLIP, CONCH, UNI, VIRCHOW2, GigaPath, H-Optimus-0, MIDNIGHT) on six tile-level
datasets at six label fractions (1\% to 100\%) with three seeds, 864 linear
probe runs in total, extended with a five-hospital Camelyon17 study of 20
directed transfer pairs that holds the label definition fixed. We report AUROC,
Expected Calibration Error before and after temperature scaling, Brier score,
and a Composite Reliability Index (CRI). Four phenomena emerge.
(i)~Under full supervision AUROC rank predicts post-temperature calibration
rank consistently across datasets (pooled within-dataset Spearman
$\rho = 0.52$, between-dataset heterogeneity $I^2 = 0\%$), whereas at 1\%
labels the relationship is no longer detectable and becomes dataset-specific
($\rho = -0.24$, $I^2 = 67\%$), a significant difference ($\Delta\rho = 0.62$,
95\% CI $[0.08, 1.13]$). (ii)~On the harder datasets the linear probe collapses
to a single predicted class below a dataset-specific label fraction.
(iii)~Whether additional source-domain labels help or harm target calibration
tracks how well the probe transfers rather than the presence of shift: in our
settings, below a transfer AUROC of roughly 0.85 more labels worsen target ECE
in 86--88\% of (model, pair) combinations and above it in 1--8\%, an
association that holds within a single dataset and among combinations that
never produce degenerate predictions. (iv)~Post-hoc temperature scaling removes more than 0.4 of ECE
in some cells yet fails to reduce it in 49 of 288, the fitted
temperatures show that the probe's miscalibration is predominantly
under-confidence (252 of 288 cells), and a temperature fitted at
the source is actively harmful under failing transfer.
PRISM is released as a pip-installable package with reference results for all
288 (model, dataset, label fraction) cells, pre-computed embeddings wherever
dataset and model licences permit redistribution, and extraction scripts with
checksums elsewhere. Code and
data: \url{https://github.com/alicemkyn/prism-benchmark}.
\end{abstract}

\section{Introduction}
\label{sec:introduction}

\begin{figure}[t]
\centering
\includegraphics[width=0.85\linewidth]{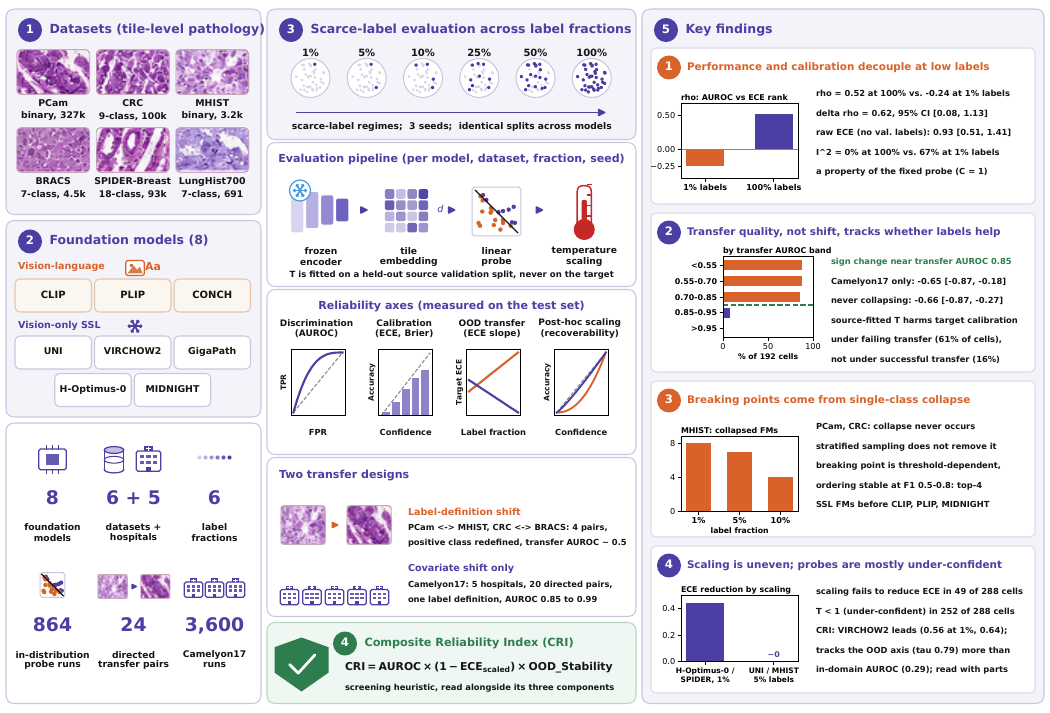}
\caption{\textbf{PRISM overview.} Eight foundation models are evaluated on
six tile-level datasets at six label fractions with three seeds (864 linear
probe runs), plus a five-hospital Camelyon17 grid of 20 directed transfer pairs
(3{,}600 runs), measuring discrimination, calibration, out-of-distribution
transfer and post-hoc recoverability. Two transfer designs separate
label-definition shift from covariate shift. Right: the four findings, with
the numbers reported in Section~\ref{sec:experiments} (finding 3 counts FMs
whose probe collapses in all three seeds). The Composite Reliability
Index is a screening heuristic read alongside its three components.}
\label{fig:1_overview}
\end{figure}

Pathology foundation models (FMs) trained with self-supervised learning (SSL),
such as UNI~\cite{chen2024uni}, CONCH~\cite{lu2024conch},
VIRCHOW2~\cite{zimmermann2024virchow2} and GigaPath~\cite{xu2024gigapath},
now set the state of the art on tile-level classification. The dominant
evaluation paradigm reports AUROC or accuracy at full label availability on
in-distribution test sets. This has driven rapid progress in discrimination
but obscures a question central to clinical use:
\emph{can the model say ``I don't know'' when it should not be confident?}
Regulatory frameworks make the requirement explicit. The FDA Good Machine
Learning Practice principles~\cite{fda2021gmlp} call for evaluation at
clinically relevant operating points, which in practice means calibrated
probabilities, and the EU AI Act Article~13~\cite{euaiact2024} requires
high-risk systems to report accuracy alongside known limitations under
distribution shift. Neither is satisfied by an AUROC reported at 100\% labels
on a single in-distribution split.

Existing benchmarks address subsets of this requirement but not their joint
interaction (Table~\ref{tab:related}, Appendix~\ref{app:benchmarks}): THUNDER~\cite{marza2026thunder}
reports calibration but varies supervision only through fixed few-shot counts
and treats robustness adversarially; HistoVL~\cite{majzoub2025histovl} and
Thiringer et al.~\cite{thiringer2026} report calibration at full supervision
only; others report discrimination
alone~\cite{neidlinger2025, ma2025pathbench, lee2025adaptation}. No prior
study evaluates the joint interaction of FM choice, label fraction, OOD shift
and calibration, the four axes along which we take clinical reliability to be
constrained. PRISM closes this gap.

Our contributions are: (i) a unified evaluation of 8 pathology FMs across 6
datasets at 6 label fractions with 3 seeds (864 linear probe runs), extended
with a five-hospital Camelyon17 grid of 20 directed transfer pairs (3{,}600
further runs); (ii) a Composite Reliability Index (CRI) that multiplies
AUROC, calibration and OOD stability, reported alongside its components;
(iii) an analysis showing that whether additional source labels help or harm
target calibration is associated with transfer quality, with a sign change near
transfer AUROC 0.85 in the evaluated settings; and (iv) an open-source, pip-installable package
(\texttt{prism\_bench}) with licence-permitted pre-computed embeddings and a
\texttt{PRISMEvaluator} API for evaluating new models against the eight
reference FMs without re-running embedding extraction.

Across these runs PRISM surfaces four phenomena, each a property of this
evaluation pipeline: AUROC and calibration rankings come apart as supervision
shrinks; the probe shows dataset-specific breaking points produced by
single-class collapse; more source labels worsen target calibration on
label-definition transfer pairs and improve it across Camelyon17 hospitals,
with transfer AUROC tracking the sign across both designs; and temperature
scaling is far from uniformly effective, harmful when fitted at a source from
which the probe does not transfer.

\section{Related work}
\label{sec:related}

Pathology FMs span vision-only DINOv2-style encoders~\cite{oquab2023dinov2}
and vision-language variants, from 86M to 1.1B parameters and from 92k to 3.1M
pre-training WSIs (Table~\ref{tab:models}); recent surveys cover the design
space~\cite{li2025survey}.
Performance-focused benchmarks compare many models on discrimination
alone~\cite{wolflein2023benchmarking, kang2023benchmarking, gatopoulos2024eva,
zhang2025pathobench, neidlinger2025, ma2025pathbench, lee2025adaptation,
breen2025ovarian, campanella2025clinical}, and Alfasly
et al.~\cite{alfasly2024fanfare} argue that such evaluation is underpowered
relative to clinical claims. Calibration-focused benchmarks are rarer. THUNDER~\cite{marza2026thunder} is the
most comprehensive, reporting ECE and its variants for 23 FMs with few-shot and
adversarial evaluations; HistoVL~\cite{majzoub2025histovl} reports ECE for four
vision-language models; Nunes et al.~\cite{nunes2025beyond, nunes2024mil} argue
for calibration as a primary CPath metric; and Thiringer et al.~\cite{thiringer2026}
evaluate 14 FMs under scanner shift with ECE and Brier at full supervision.
Neither the decoupling of discrimination from calibration nor the degradation
of uncertainty under shift is new: Ovadia et al.~\cite{ovadia2019trust}
establish both outside pathology and Thagaard et al.~\cite{thagaard2020trust}
the second within it, both at full supervision. What PRISM adds is the label-fraction axis for a
generation of FMs that postdates these studies, both raw and post-hoc-recovered
calibration, and the direction of transfer, evaluated jointly with FM choice.
Temperature scaling~\cite{guo2017calibration} and ECE~\cite{nixon2019measuring} are used as standard.

\section{Measuring reliability of pathology foundation models}
\label{sec:method}

\subsection{Models and datasets}
\label{sec:models_datasets}

\paragraph{Foundation models.} PRISM evaluates eight FMs selected to span the
design space of pathology pre-training (Table~\ref{tab:models}): three
vision-language contrastive models (CLIP~\cite{radford2021clip} as a
non-pathology baseline, PLIP~\cite{huang2023plip},
CONCH~\cite{lu2024conch}) and five vision-only SSL models
(UNI~\cite{chen2024uni}, GigaPath~\cite{xu2024gigapath},
H-Optimus-0~\cite{bioptimus2024hoptimus},
MIDNIGHT~\cite{karasikov2025midnight}, VIRCHOW2~\cite{zimmermann2024virchow2}),
from 86M to 1.1B parameters and from 92k to 3.1M pre-training WSIs.
Preprocessing is not uniform across the eight FMs and we state what each
received (Appendix~\ref{app:models}); embeddings are L2-normalized. Each FM
is embedded once under one pipeline held fixed across every label fraction
and transfer pair, so comparisons \emph{within} a model are unaffected by this
heterogeneity, while comparisons \emph{between} models at a fixed fraction are
confounded by it. In particular CLIP, PLIP and MIDNIGHT received ImageNet
statistics rather than their documented normalization. A partial check:
Camelyon17, embedded with every FM's documented transform, shares PCam's task.
The AUROC rankings of the eight FMs on the two agree (Kendall $\tau$ 0.50--0.79
across fractions; CLIP and PLIP last on both), whereas the scaled-ECE rankings
do not ($\tau$ between $-0.50$ and $0.29$). This bounds rather than isolates
the preprocessing effect, since the datasets also differ in source and tile
size.

\paragraph{Datasets.} Six tile-level datasets (Table~\ref{tab:datasets})
span task type (binary PCam~\cite{veeling2018pcam} and
MHIST~\cite{wei2021mhist}; 7-class BRACS~\cite{brancati2022bracs} and
LungHist700~\cite{diosdado2024lunghist700}, 9-class CRC~\cite{kather2019crc},
18-class SPIDER-Breast~\cite{nechaev2025spider}), scale (691 to 327{,}680
tiles; absolute training counts per fraction are in
Table~\ref{tab:label_budget}), and organ (lymph node, colorectal, breast,
lung). The four-pair OOD evaluation is built from PCam$\leftrightarrow$MHIST
and CRC$\leftrightarrow$BRACS, cross-institution pairs that can be assembled
from publicly available data.

\begin{table}[h]
\caption{Tile-level datasets included in PRISM, ordered by sample count: six
datasets spanning binary to 18-class classification, with counts varying by
nearly three orders of magnitude. CRC is NCT-CRC-HE-100K, split 70/15/15.
RUA: research-use agreement without redistribution rights.}
\label{tab:datasets}
\centering
\small
\begin{tabular}{llrrll}
\toprule
Dataset & Task & Classes & Samples & License & Source \\
\midrule
PCam          & Binary (tumor)   & 2  & 327{,}680 & CC0          & Veeling et al.~\cite{veeling2018pcam} \\
CRC           & Multiclass       & 9  & 100{,}000 & CC-BY 4.0    & Kather et al.~\cite{kather2019crc} \\
MHIST         & Binary (HP/SSA)  & 2  &   3{,}152 & RUA          & Wei et al.~\cite{wei2021mhist} \\
BRACS         & Multiclass       & 7  &   4{,}539 & CC-BY-NC 4.0 & Brancati et al.~\cite{brancati2022bracs} \\
SPIDER-Breast & Multiclass       & 18 &  92{,}900 & CC-BY-NC 4.0 & Nechaev et al.~\cite{nechaev2025spider} \\
LungHist700   & Multiclass       & 7  &      691  & CC-BY 4.0    & Diosdado et al.~\cite{diosdado2024lunghist700} \\
\bottomrule
\end{tabular}
\end{table}

\subsection{Evaluation protocols}
\label{sec:protocols}

\paragraph{Label-fraction sweep.} On frozen embeddings ($d \in [512, 2560]$)
we train an $\ell_2$-regularized logistic regression probe (scikit-learn LBFGS,
$C = 1.0$) on stratified training subsets at label fractions
$\{0.01, 0.05, 0.10, 0.25, 0.50, 1.00\}$ with three seeds (42, 123, 456);
splits are identical across models by construction and test sets are held
fixed per dataset, for $8 \times 6 \times 6 \times 3 = 864$ runs.

\paragraph{Scope of the conclusions.} Every result is a property of one
evaluation pipeline: frozen tile embeddings, a logistic probe whose
regularization is held at $C = 1$ from 7 to 262{,}144 training tiles, and
scalar temperature scaling. Holding $C$ fixed keeps every cell under an identical protocol, but $C$
shapes calibration: across a five-value sweep the AUROC ranking of FMs is stable (mean Kendall $\tau = 0.82$ against $C = 1$)
while the raw-ECE ranking is not ($\tau = 0.02$). On the three datasets
swept (MHIST, LungHist700, BRACS; raw ECE), the 1\%-versus-100\% decoupling
contrast of Section~\ref{sec:experiments} is pronounced at $C = 1$
($\Delta\rho = 0.93$), weaker at $C = 10$ and $100$ (0.06 and 0.37) and
reversed at $C \leq 0.1$ (Appendix~\ref{app:probe}).
Calibration comparisons between FMs are therefore properties of this
pipeline, not intrinsic properties of the FMs.

\paragraph{Calibration with post-hoc recoverability.} For each linear probe we
report Expected Calibration Error (ECE)~\cite{guo2017calibration} both before
and after temperature scaling. Partitioning the $N$ test predictions into bins
$B_1,\dots,B_M$ by confidence,
\begin{equation*}
\mathrm{ECE} = \sum_{m=1}^{M} \frac{|B_m|}{N}
  \bigl| \mathrm{acc}(B_m) - \mathrm{conf}(B_m) \bigr|,
\qquad
\mathrm{Brier} = \frac{1}{N}\sum_{i=1}^{N} (\hat{p}_i - y_i)^2 ,
\end{equation*}
where $\mathrm{conf}(B_m)$ is the mean confidence within bin $m$ and
$\mathrm{acc}(B_m)$ the corresponding empirical frequency. On the four
multiclass tasks confidence is the top-label probability and
$\mathrm{acc}(B_m)$ the top-label accuracy; on the two binary tasks confidence
is the predicted probability $\hat{p}_i$ of the positive class and
$\mathrm{acc}(B_m)$ the observed positive rate, so ECE measures positive-class
calibration. We use $M = 15$ equal-width bins; equal-mass bins leave the
decoupling contrast unchanged (Appendix~\ref{app:probe}). The Brier score above is the binary
form; on multiclass tasks we report its one-vs-rest average over classes.
\emph{Raw} ECE is computed on the probe's own outputs;
\emph{scaled} ECE after dividing the logits by a scalar temperature $T$ fitted
by negative log-likelihood on a held-out validation split ($T \in [0.1, 10]$),
so the difference isolates miscalibration of scale from miscalibration of
ordering. For binary tasks the temperature is fitted on the two-column logit
$[-z, z]$ of the log-odds $z$; we report it divided by two so that $T = 1$
means the probe's own probabilities are kept. For transfer evaluation $T$ is fit on the \emph{source} validation
split and applied to the target without target labels. Brier score is reported
as a strictly proper scoring rule. The recoverability axis, raw minus scaled
ECE, is itself an evaluation target: a single scalar parameter is the cheapest
safeguard a deployment team can afford.

\paragraph{Camelyon17: transfer with the label definition held fixed.}
The four transfer pairs below vary the label definition together with the
acquisition conditions. To separate the two we add
Camelyon17~\cite{bandi2018camelyon17} (WILDS release~\cite{koh2021wilds}):
patches from \textbf{five hospitals} annotated for one binary task (tumour
against normal in lymph node) under a single label definition, differing only
in scanner, staining and patient population. Treating each hospital as a
domain gives \textbf{20 directed transfer pairs}. We sample 20{,}000 patches
per hospital, label-stratified at 50/50, split 70/15/15 at patch level, since
a slide-disjoint split would leave tumour prevalence between 0.1\% and 83\%
across partitions and confound ECE with prevalence; leakage is impossible in
the transfer grid, where source and target are different hospitals, and only
the in-distribution numbers are optimistic (reported as an upper bound). Here
every FM is embedded under \emph{its own} documented transform
(Appendix~\ref{app:camelyon-details}), so the cross-model comparison is
controlled.

\paragraph{Cross-domain OOD pairs.} Beyond in-distribution evaluation we
construct four cross-domain transfer pairs:
PCam$\to$MHIST, MHIST$\to$PCam, CRC$\to$BRACS, BRACS$\to$CRC. For each pair
we train a linear probe on the source at every label fraction and evaluate
AUROC, ECE and Brier on the target test split under the binary label mapping
of Table~\ref{tab:oodmap} (Appendix~\ref{app:additional}). These pairs are not clean covariate shift: tissue,
acquisition site and the definition of the positive class all change together,
so they measure whether a probe trained on one binary pathology task retains
calibrated signal on another.

\begin{table}[t]
\centering
\small
\caption{Absolute label budget. Training labels $n_{\text{train}}$ at each label fraction and the size of the validation split used for temperature scaling (and CRI). The validation split is held at its full size at every fraction, so at 1\% it accounts for 90--96\% of all labels used; raw ECE uses no validation labels. Camelyon17 counts are per source hospital.}
\label{tab:label_budget}
\setlength{\tabcolsep}{4pt}
\begin{tabular}{lrrrrrrr}
\toprule
& & \multicolumn{4}{c}{$n_{\text{train}}$ at label fraction} & & \\
\cmidrule(lr){3-6}
\textbf{Dataset} & $K$ & 1\% & 5\% & 10\% & 100\% & $n_{\text{val}}$ & val.\ share @1\% \\
\midrule
PCam & 2 & 2{,}620 & 13{,}106 & 26{,}214 & 262{,}144 & 32{,}768 & 93\% \\
BRACS & 7 & 33 & 180 & 362 & 3{,}657 & 312 & 90\% \\
CRC & 9 & 696 & 3{,}495 & 6{,}996 & 70{,}000 & 15{,}000 & 96\% \\
MHIST & 2 & 18 & 92 & 184 & 1{,}849 & 326 & 95\% \\
LungHist700 & 7 & 7 & 22 & 45 & 483 & 103 & 94\% \\
SPIDER-Breast & 18 & 682 & 3{,}431 & 6{,}868 & 68{,}730 & 12{,}128 & 95\% \\
Camelyon17$^\dagger$ & 2 & 140 & 700 & 1{,}400 & 14{,}000 & 3{,}000 & 96\% \\
\bottomrule
\end{tabular}
\par\smallskip{\footnotesize $^\dagger$Per hospital; each of the five hospitals contributes the same counts.}
\end{table}

\paragraph{What a label fraction consumes.} A fraction subsamples the training
split only, so the same percentage means very different amounts of data:
7 training tiles on LungHist700 and 2{,}620 on PCam at 1\%
(Table~\ref{tab:label_budget}). The validation split used for temperature
fitting is held at full size and supplies 90--96\% of all labels a cell
consumes at 1\%. We therefore separate two kinds of low-label calibration
claim. \emph{Raw} ECE uses no validation labels and reflects the training
budget alone; \emph{scaled} ECE at 1\% reflects that budget plus a full
validation split. A proportional protocol that subsamples validation by the
same fraction moves scaled ECE at 1\% by 0.009 and preserves the FM ranking
(Appendix~\ref{app:full_results}), but leaves as few as three validation
samples on MHIST; low-label calibration claims are therefore stated for raw
ECE alongside scaled ECE.

\paragraph{Composite Reliability Index.} The three reliability axes can pull in
opposite directions: an FM may have strong AUROC but poor calibration, or
strong in-distribution behavior but catastrophic OOD transfer. To summarize
joint reliability we define the Composite Reliability Index:
\begin{equation}
\mathrm{CRI} = \mathrm{AUROC} \times (1 - \mathrm{ECE}_{\text{scaled}}) \times \mathrm{OOD\_Stability}
\label{eq:cri}
\end{equation}
where $\mathrm{ECE}_{\text{scaled}} \in [0, 1]$ is the post-temperature ECE
(clipped to $[0,1]$; no further normalization is applied), and
$\mathrm{OOD\_Stability} \in [0, 1]$ is the ratio of OOD AUROC to in-distribution
AUROC averaged across the relevant transfer pairs (clipped to 1 from above).
For the CRC and BRACS source pairs the denominator is the source's multiclass
macro AUROC and the numerator a binary AUROC under the mapping of
Table~\ref{tab:oodmap}.
The multiplicative form is deliberate: a model that scores zero on any single
axis cannot achieve a high CRI, regardless of how strong it is on the others.
This penalty structure reflects the view that an FM should be acceptable on
\emph{all} axes, not just the one a benchmark happens to emphasize.

Two properties of the index should be kept in mind. First,
$\mathrm{OOD\_Stability}$ is a ratio, so it partially rewards models whose
in-distribution performance was already weak: the composite tracks the OOD
axis (Kendall $\tau = 0.79$ against the OOD AUROC ranking) far more closely
than the in-distribution axis ($\tau = 0.29$), so we report the three axes
separately and treat the composite as a summary of them. (An earlier release
computed $1 - \mathrm{CV}$ of in-distribution AUROC in place of the ratio; all
values here use Equation~\ref{eq:cri}.) Second, the calibration term inherits
the probe dependence described above, so CRI is meaningful only under a fixed
probe protocol. The combination rule is not what drives it (multiplicative,
arithmetic, geometric and worst-axis aggregations agree at $\tau$ between 0.71
and 1.00). CRI is a hand-specified screening heuristic for down-selection; it
is not validated against any clinical outcome and does not measure clinical
readiness.

\paragraph{Accessibility.}\label{sec:design}
PRISM ships as a pip-installable package, \texttt{prism\_bench}, with a
\texttt{PRISMEvaluator} class, a CLI and the reference results. Embeddings
are extracted once (about 15~GPU-hours on A100 hardware), after which all 864
probe runs complete in under 30~minutes on one CPU. Licences determine what
can be redistributed: embeddings are released for five of the six datasets
(all but MHIST, whose research-use agreement forbids redistribution) and for
Camelyon17, from the five FMs whose licences permit sharing model outputs
(CLIP, PLIP, GigaPath, H-Optimus-0, MIDNIGHT); for MHIST and for UNI, CONCH
and VIRCHOW2 we release extraction scripts, split indices and SHA256
checksums so that licensed users can regenerate and verify them.

\section{Experiments}
\label{sec:experiments}

We present results across the four reliability axes; per-cell tables,
bootstrap intervals and supporting figures are in the appendix.

\begin{figure}[t]
\centering
\includegraphics[width=0.64\linewidth]{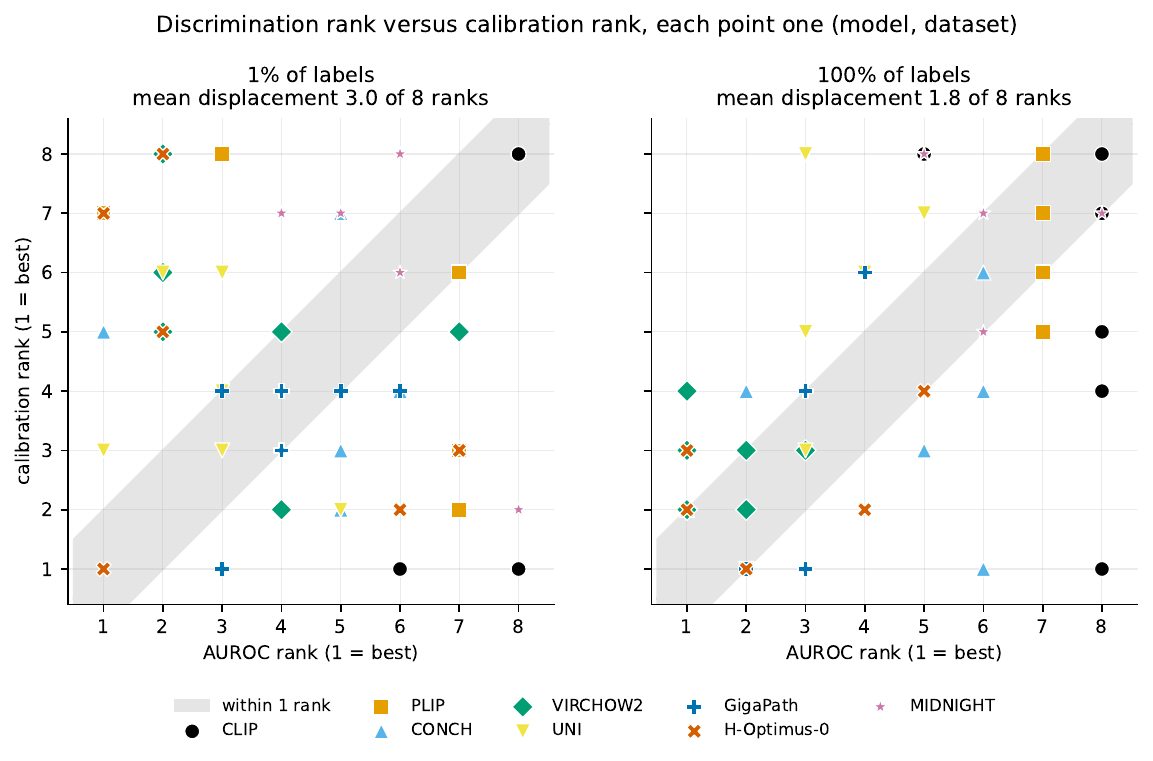}
\caption{\textbf{Discrimination rank against calibration rank.} Each point is
one (model, dataset) cell: its position on the horizontal axis is the model's
AUROC rank within that dataset, on the vertical axis its rank by
post-temperature ECE, both with 1 best. Points on the diagonal are models
whose two ranks agree and the shaded band marks agreement to within one place.
Mean displacement is 3.0 of 8 positions at 1\% labels against 1.8 under full
supervision. The largest discords at 1\% labels are CLIP on BRACS, LungHist700
and PCam (last on AUROC, first on calibration) and H-Optimus-0 on PCam (first
on AUROC, seventh on calibration).}
\label{fig:7_decoupling}
\end{figure}

\begin{figure}[t]
\centering
\includegraphics[width=0.62\linewidth]{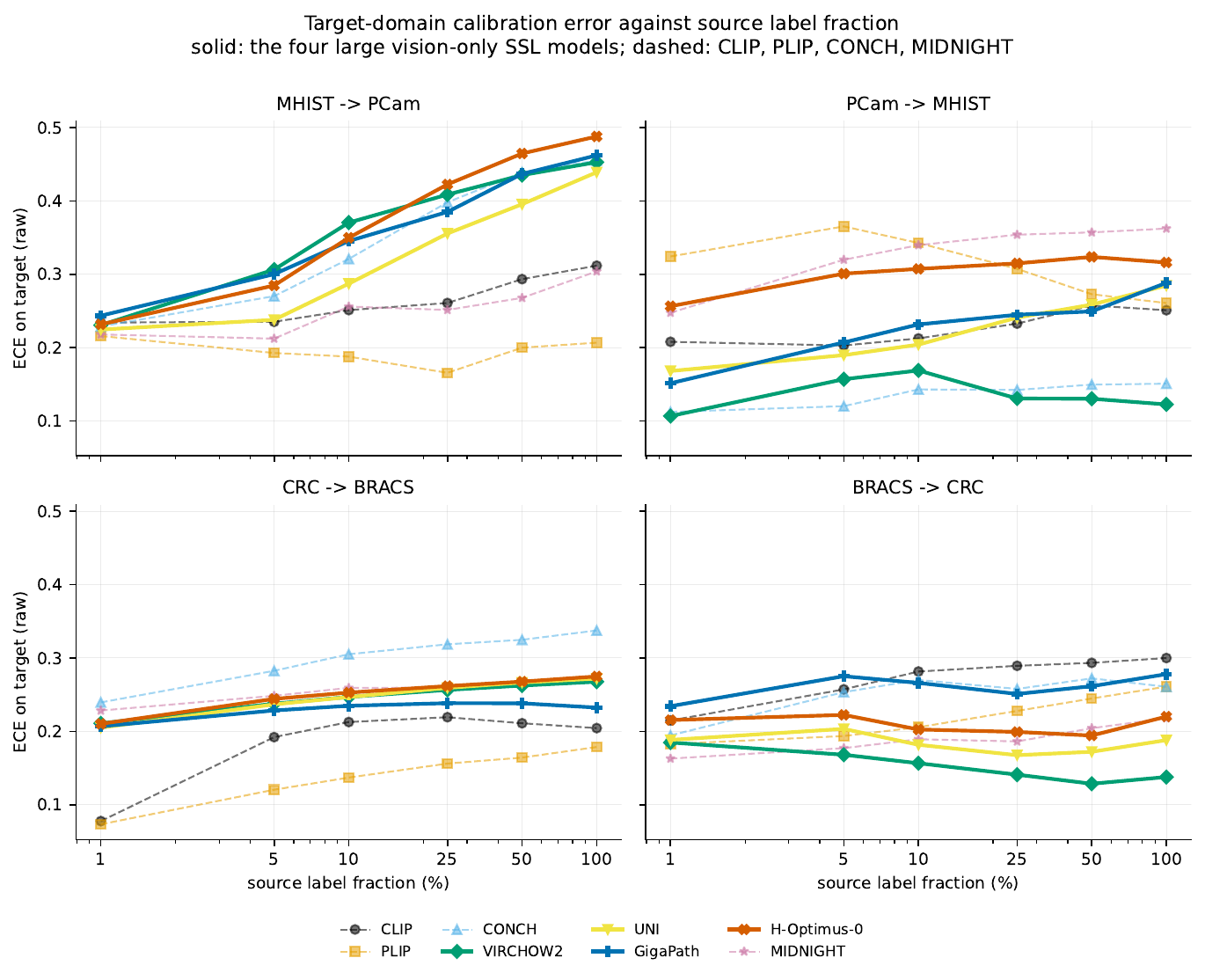}
\caption{\textbf{Reverse OOD calibration scaling.} Raw ECE on the transfer
target against source-domain label fraction, at all six fractions rather than
the two endpoints. On MHIST$\to$PCam the four large vision-only SSL models (UNI,
VIRCHOW2, GigaPath, H-Optimus-0; solid lines) rise from roughly 0.23 to
between 0.44 and 0.49, monotonically in the label fraction and consistently
across seeds; CONCH (dashed) rises as strongly, CLIP and MIDNIGHT more weakly,
and PLIP stays flat. The three remaining pairs show
the effect weakly or not at all, so the phenomenon is pair-dependent and we do
not claim it holds for transfer in general.}
\label{fig:10_ood_ece}
\end{figure}

\paragraph{Performance-reliability decoupling.} Figure~\ref{fig:7_decoupling}
contrasts AUROC and post-temperature ECE rankings across the six datasets.
Pooling within-dataset ranks across the 48 (model, dataset) cells, the
correlation between AUROC rank and calibration rank is $\rho = -0.24$ at 1\%
labels (95\% CI $[-0.54, 0.36]$) against $\rho = 0.52$ at full supervision
($[0.27, 0.72]$), and the contrast is significant, $\Delta\rho = 0.62$
($[0.08, 1.13]$), under a cluster bootstrap that resamples whole datasets and
one seed per cell (seed-averaging gives $\Delta\rho = 0.75$; we quote the
conservative value). A random-effects meta-analysis over the six per-dataset
correlations agrees ($\rho = -0.30$ and $0.57$) and adds that between-dataset
heterogeneity is $I^2 = 0\%$ at full supervision but $67\%$ at 1\% labels.
Because scaled ECE at 1\% borrows a full validation split
(Table~\ref{tab:label_budget}), we repeat the analysis on raw ECE, which uses
no validation labels: the contrast is larger, $\rho = -0.46$ $[-0.85, -0.05]$
at 1\% against $0.47$ $[0.03, 0.86]$ at full supervision,
$\Delta\rho = 0.93$ $[0.51, 1.41]$, and remains significant at 5\% and 10\%
labels (0.52 and 0.37). It is, however, specific to the probe's
regularization (Section~\ref{sec:protocols}). A model's calibration rank sits 3.0 of 8 positions from its AUROC rank
at 1\% labels against 1.8 at full supervision. The largest discords run in
both directions. CLIP ranks last on AUROC yet first on post-temperature ECE on
three datasets (BRACS, LungHist700, PCam); on BRACS its probe assigns 92\% of
the test set to one class (macro-F1 0.08), so it is well calibrated largely
because it commits to almost nothing. Conversely, H-Optimus-0 has the top
AUROC on PCam at 1\% labels (0.982) but ranks seventh of eight on calibration.
A practitioner selecting on AUROC alone would therefore not select the
best-calibrated model, and selecting on ECE alone would favour a near-degenerate
one. The correlation is negative at 1\% labels within each seed taken
separately, so the pattern is not an averaging artefact
(Appendix~\ref{app:full_results}).

\paragraph{Breaking points.} Macro-F1 against label fraction for every
(model, dataset) pair is given in Figure~\ref{fig:4_breaking_point}
(Appendix~\ref{app:additional}). PCam and CRC are close to saturation in AUROC
already at 1\% labels (0.88--0.98 and 0.98--1.00), although macro-F1 there
still ranges from 0.80 (CLIP) to 0.94 and 0.99; BRACS, MHIST and LungHist700
reveal large label-efficiency differences. CLIP and PLIP never reach
F1 = 0.5 on BRACS, while UNI and VIRCHOW2 cross it on every dataset by 25\%
labels and GigaPath and H-Optimus-0 by 50\%. At 10\% labels only four FMs
(CONCH, GigaPath, UNI, VIRCHOW2) reach the threshold on BRACS and none does on
MHIST; we state this as a property of the benchmark tasks, not a deployment
criterion. The breaking point is threshold-dependent: sweeping the macro-F1
threshold from 0.4 to 0.8 moves it by 0.23 on average on BRACS and
SPIDER-Breast and by 0.64 and 0.72 on MHIST and LungHist700. The
\emph{ordering} (VIRCHOW2, UNI, GigaPath and H-Optimus-0 no later than CLIP,
PLIP and MIDNIGHT) holds at every threshold from 0.5 to 0.8, but not at 0.4 on
LungHist700, where PLIP and CONCH cross first at 1\% labels. A collapse audit
(Figure~\ref{fig:18_degeneracy}) shows why: on MHIST every FM assigns more
than 99\% of the test set to one class at 1\% labels, on PCam and CRC never,
and stratified sampling does not remove the collapse. Post-hoc recoverability
is likewise non-monotone, from removing 0.45 of ECE (H-Optimus-0 on
SPIDER-Breast at 1\% labels, 0.465 to 0.020) to none (UNI on MHIST at 5\% and 10\% labels, where
scaling slightly increases ECE); across all 288 cells scaling fails to reduce
ECE in 49, concentrated on MHIST, PCam and LungHist700. The plateau between
5\% and 25\% coincides with the probe emerging from collapse
(Figure~\ref{fig:5_recoverability}). The fitted temperatures also give the
\emph{direction} of miscalibration, which ECE alone does not. Expressed on the
probe's own logit scale, $T < 1$ (under-confidence) holds in 189 of the 192
multiclass cells ($T$ between 0.28 and 0.40 on SPIDER-Breast at 1\% labels)
and in 63 of the 96 binary cells; over-confidence ($T > 1$) appears mainly on
PCam from 25\% labels and on MHIST at 1--5\%, where the probe is collapsed.
The large raw ECE of the multiclass probes at low label fractions is therefore
under-confidence of a strongly regularized linear probe rather than
over-confidence, which is also why a single scalar removes it so readily.

\paragraph{Out-of-distribution reliability and reverse calibration scaling.}
On the four cross-domain pairs of the original design, \emph{more source-domain
labels worsen target calibration}. Figure~\ref{fig:10_ood_ece} shows OOD ECE
against source label fraction. On MHIST$\to$PCam all four large vision-only SSL
models rise monotonically and consistently across seeds: UNI from 0.225 to
0.439, VIRCHOW2 from 0.231 to 0.453, GigaPath from 0.243 to 0.462 and
H-Optimus-0 from 0.232 to 0.488. CONCH rises as strongly (0.229 to 0.455),
CLIP and MIDNIGHT more weakly (to 0.312 and 0.304), and only PLIP stays flat. Paired
with OOD AUROC the effect is severe: at 100\% labels GigaPath reaches OOD AUROC
0.487 yet OOD ECE 0.462, an order of magnitude worse than its in-distribution
ECE of 0.042. The three remaining pairs show the effect weakly or not at all,
so it is pair-dependent.

\begin{figure}[b]
\centering
\begin{minipage}[t]{0.40\linewidth}\vspace{0pt}\centering
\includegraphics[height=1.82in]{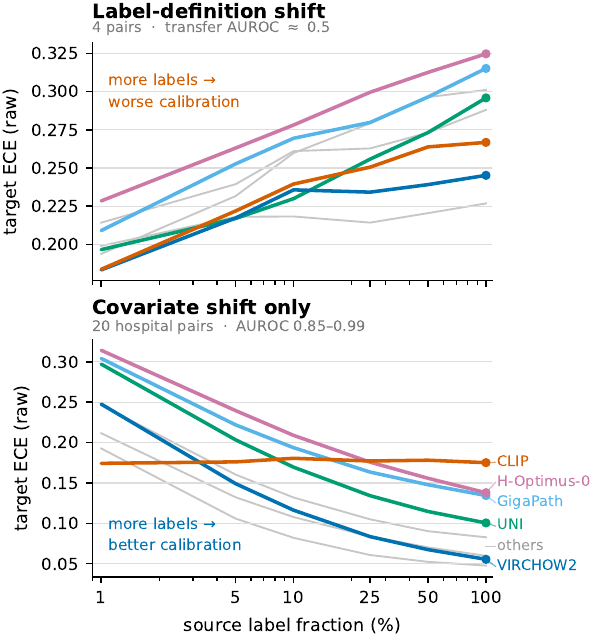}\\[-2pt]
{\small (a)}
\end{minipage}\hfill
\begin{minipage}[t]{0.56\linewidth}\vspace{0pt}\centering
\includegraphics[height=1.82in]{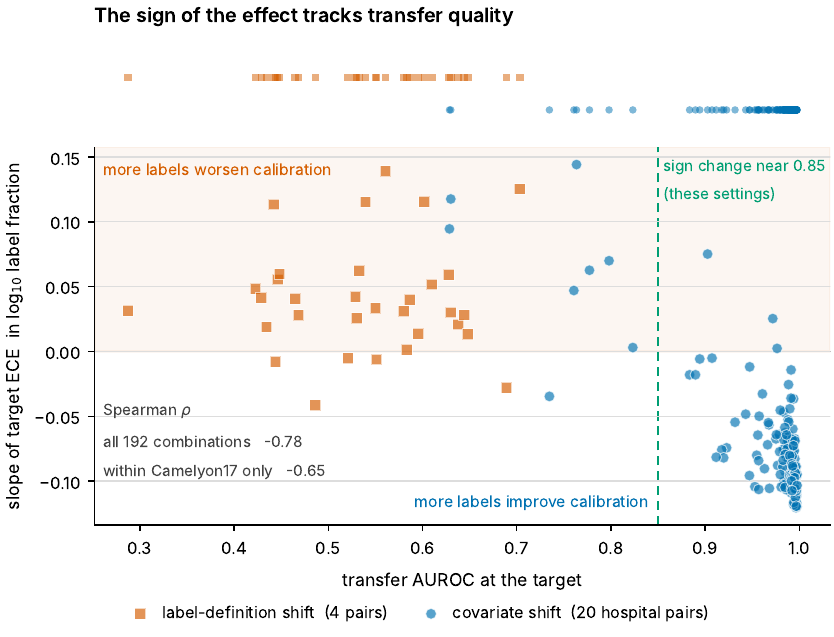}\\[-2pt]
{\small (b)}
\end{minipage}
\caption{\textbf{Reverse calibration scaling tracks transfer quality, not the
kind of shift.} (a)~The same protocol, opposite directions: raw target ECE
against source label fraction, averaged over pairs within each design. Top:
the four label-definition pairs (transfer AUROC near 0.5); bottom: twenty
Camelyon17 hospital pairs with one label definition. (b)~Each point is one
(model, pair) combination; the vertical axis is the slope of target ECE in
$\log_{10}$ label fraction, so points in the shaded region are combinations
where additional source labels worsen calibration. The strip above marks where
each design sits on the AUROC axis; the two designs overlap through the
transition zone.}
\label{fig:contrast}
\end{figure}

This is the phenomenon, not yet an explanation: the four pairs that produce it
change the label definition as well as the acquisition conditions, and a probe
trained to separate tumour from normal lymph node and evaluated against SSA
versus HP polyp labels is not performing standard OOD generalization.
Section~\ref{sec:camelyon} repeats the protocol on Camelyon17, where only
scanner, staining and patient population differ. Across all twenty hospital
pairs the effect \emph{reverses}: target ECE rises in 10 of 160 (model, pair)
combinations rather than 28 of 32, and the mean change is $-0.149$ rather than
$+0.082$ (Figure~\ref{fig:contrast}a).

\paragraph{Reverse scaling accompanies failing transfer.} The two designs differ in
transfer difficulty as well as in kind of shift (label-definition pairs
transfer at AUROC near 0.5, hospital pairs at 0.85 to 0.996), so we test
transfer quality directly over all 192 (model, pair) combinations, using the
slope of target ECE against $\log_{10}$ label fraction
(Figure~\ref{fig:contrast}b).

\textbf{Transfer quality predicts the slope, and the relation survives inside a
single dataset.} Combinations share models, hospitals and test sets, so we
attach 95\% CIs from a two-way cluster bootstrap that resamples models and
hospitals (and the four label-definition pairs) rather than combinations.
Pooling both designs, the Spearman correlation between OOD AUROC and the ECE
slope is $-0.78$ $[-0.91, -0.59]$. Within Camelyon17 alone, where dataset, task
and kind of shift are held constant, it is $-0.65$ $[-0.87, -0.18]$ ($-0.91$
across the eight FMs and $-0.75$ across the twenty pairs); the dataset is not
what drives the contrast.

\textbf{In these settings the sign changes near AUROC 0.85.} Binning all 192
combinations by transfer AUROC, ECE rises in 88\%, 88\% and 86\% of
combinations below 0.55, between 0.55 and 0.70, and between 0.70 and 0.85
respectively, then in 8\% between 0.85 and 0.95 and 1\% above 0.95; the
difference in rising rate below and above 0.85 is 86 percentage points
$[66, 100]$. The two transition bands contain
combinations from \emph{both} designs, so covariate-shift transfers that
perform poorly also show the effect.

\textbf{It is not an artefact of degenerate prediction.} Single-class collapse
occurs in 57\% of label-definition cells and 0.7\% of hospital cells, which
makes it a candidate explanation. Restricting to the 163 of 192 combinations
that never collapse, transfer AUROC still predicts the slope at $-0.66$
$[-0.87, -0.27]$, and in a joint regression it accounts for $R^2 = 0.65$ against 0.44 for
degeneracy alone.

\textbf{The clearest single case is CLIP.} On Camelyon17, with dataset, task
and shift type fixed, CLIP transfers at a mean AUROC of 0.849 (over pairs and
fractions) and shows rising ECE in 8 of its 20 pairs; its 8 pairs below 0.85
are the only Camelyon17 ones, so the within-Camelyon17 rate below the
transition rests on CLIP alone. The other seven FMs transfer at 0.966 to 0.995
and show it in 2 of 140. In the evaluated settings, then, \emph{adding source
labels worsened target calibration when the probe failed to transfer and
improved it when the probe transferred}. This is an association, not a causal
identification, and the transition point is specific to this probe, these
datasets and these pairs; label-definition shift is simply the most reliable
way to push a probe below it. As a heuristic diagnostic, calibration that
degrades as source labels are added indicated a non-transferring probe.

Post-hoc correction does not rescue the failing case. Fitting the temperature
on a held-out \emph{source} validation split \emph{increases} target ECE for
all eight FMs on CRC$\to$BRACS from 5\% source labels and on MHIST$\to$PCam
from 25\%, and worsens ECE in 61\% of cells across the four original pairs; on the hospital pairs, where transfer
succeeds, it worsens ECE in only 16\%.

\paragraph{Aggregated reliability via the Composite Reliability Index.}
Figure~\ref{fig:6_cri} (Appendix~\ref{app:additional}) reports CRI per (model,
label fraction) cell. VIRCHOW2 leads in both regimes (0.642 at full
supervision, 0.564 at 1\% labels), with best-to-worst spreads of 0.17 and
0.15. Two placements show what the index rewards: UNI ranks last at full
supervision despite in-distribution AUROC 0.954 because its transfer AUROC
(0.49) gives it the lowest $\mathrm{OOD\_Stability}$ (0.52), whereas CLIP,
last on in-distribution AUROC on five of six datasets, places fourth and
second because a ratio term rewards a model with little to lose under
transfer. Section~\ref{sec:camelyon} returns to a transfer design in which
$\mathrm{OOD\_Stability}$ measures what it was defined to measure.

\subsection{Transfer under covariate shift alone}
\label{sec:camelyon}

Camelyon17 supplies 20 directed hospital pairs sharing one label definition,
run under the identical protocol: 720 in-distribution and 2{,}880 transfer
runs. Transfer is meaningful here in a
way the original pairs could not guarantee: at full supervision the eight FMs
reach mean OOD AUROC between 0.885 (CLIP) and 0.996 (VIRCHOW2) against roughly
0.50 on the label-definition pairs, and single-class collapse occurs in 0.7\%
of cells against 57\% there.

Under covariate shift alone additional source labels \emph{improve} target
calibration (Figure~\ref{fig:contrast}a): for UNI, VIRCHOW2, GigaPath and
H-Optimus-0, ECE rises in 1 of 80 combinations with a mean change of $-0.184$,
and a source-fitted temperature worsens target ECE in 16\% of cells against
61\% on the original pairs. The reliability index also behaves as intended
once label semantics are fixed: on same-task transfer the CRI ordering places
the pathology-specific FMs (VIRCHOW2, UNI, H-Optimus-0, GigaPath, MIDNIGHT)
above CONCH, PLIP and CLIP, and the composite agrees with all three of its
components (Kendall $\tau = 0.71$, 0.93 and 0.86 against in-distribution
AUROC, OOD AUROC and the drop); the ratio term rewarded weak in-distribution
models on the original pairs only because transfer there was near chance for
every model. Averaging over models at full supervision, the
\emph{target} hospital matters more than the source for calibration (standard
deviation of per-hospital mean raw ECE 0.043 across targets against 0.020
across sources) and about equally for AUROC (0.008 against 0.008)
(Figure~\ref{fig:hospital-matrix}). Label-budget accounting for this dataset
is in Appendix~\ref{app:camelyon-details}.

\section{Conclusion}
\label{sec:conclusion}

PRISM evaluates pathology FM reliability under joint variation of model,
label fraction, calibration and transfer (864 in-distribution and 3{,}600
Camelyon17 runs). Under a fixed frozen-embedding linear-probe pipeline it
documents low-label decoupling of discrimination and calibration,
collapse-driven breaking points, a sign change in reverse calibration scaling
near transfer AUROC 0.85 in the evaluated settings, and far-from-uniform
post-hoc recoverability with predominantly under-confident probes. Calibration that degrades as source labels accumulate is a cue to measure
transfer before collecting more labels; the Composite Reliability Index is
reported alongside its components, not in place of them.

\paragraph{Limitations and broader impact.} All findings concern tile-level
classification with frozen representations under one probe protocol; they
describe model behaviour under these benchmark conditions, not readiness for
clinical deployment; slide-level evaluation, confounded with the
aggregator~\cite{shao2021transmil}, is out of
scope~\cite{gatopoulos2024eva, zhang2025pathobench}. The probe's
regularization is fixed at $C = 1$; calibration rankings and the low-label
decoupling depend on it (Section~\ref{sec:protocols}), and a
validation-tuned-$C$ replication, like a slide-disjoint Camelyon17
sensitivity analysis, is left to future work. An MLP probe over every cell preserves the direction of the decoupling and
reverse-scaling effects (Appendix~\ref{app:probe}). The 0.85 transition is an
association on two transfer designs, not a general threshold. Camelyon17 isolates covariate
but not scanner-only shift~\cite{thiringer2026}. CRI is
a screening heuristic, not a deployment criterion; we recommend institutional
validation on local data before any clinical use.

\begin{ack}
This work received no external funding. The author declares no competing
interests.
\end{ack}

\bibliographystyle{plain}
\bibliography{references}

\FloatBarrier
\clearpage
\appendix

\begin{center}
{\LARGE\bf Appendix}
\end{center}
\vspace{1em}

\section{Included foundation models}
\label{app:models}

This appendix lists architectural and training details for each of the eight
foundation models supported in PRISM, complementing Table~\ref{tab:models}.
All eight models share a Vision Transformer
backbone~\cite{dosovitskiy2020vit}, with four of the pathology-pretrained
SSL models (UNI, GigaPath, H-Optimus-0, MIDNIGHT) using the
DINOv2~\cite{oquab2023dinov2} self-supervised training recipe at varying
scales. Each model is loaded through its native HuggingFace interface, and the
preprocessing applied to each is given in Table~\ref{tab:preproc}. Five of the
eight share a resize to 224 with a centre crop and ImageNet normalization
statistics; GigaPath resizes to 256; H-Optimus-0 uses its own normalization
statistics; CONCH uses the transform returned by its own loader. All embeddings
are L2-normalized.

\begin{table}[H]
\caption{The eight foundation models evaluated. Sizes range from 86M to 1.1B parameters.}
\label{tab:models}
\centering
\small
\begin{tabular}{llrrl}
\toprule
Model & Type & Size & $d$ & Pre-training corpus \\
\midrule
CLIP        & Vision-Language & 86M  & 768   & WebImageText (general) \\
PLIP        & Vision-Language & 86M  & 768   & Twitter pathology pairs \\
CONCH       & Vision-Language & 86M  & 512   & Curated pathology pairs \\
UNI         & Vision          & 307M & 1024  & 100k WSIs (Mass-100K) \\
VIRCHOW2    & Vision          & 632M & 2560  & 3.1M WSIs \\
GigaPath    & Vision          & 1.1B & 1536  & 171k WSIs (Prov-Path) \\
H-Optimus-0 & Vision          & 1.1B & 1536  & 500k WSIs \\
MIDNIGHT    & Vision          & 1.1B & 1536  & TCGA + NKI-80k (92k WSIs) \\
\bottomrule
\end{tabular}
\end{table}

\begin{table}[h]
\caption{Preprocessing applied to each FM during feature extraction. Each model
is embedded once under one pipeline, held fixed thereafter.}
\label{tab:preproc}
\centering
\small
\begin{tabular}{llll}
\toprule
Model(s) & Resize & Crop & Normalization \\
\midrule
CLIP, PLIP, MIDNIGHT, UNI, VIRCHOW2 & 224 & 224 & ImageNet \\
GigaPath & 256 & 224 & ImageNet \\
H-Optimus-0 & 224 & 224 & model-specific \\
CONCH & \multicolumn{3}{l}{transform supplied by the model loader} \\
\bottomrule
\end{tabular}
\end{table} For vision-language
models we use only the image encoder; the text encoder is unused. Embedding
dimensionality $d$ is listed in Table~\ref{tab:models}: 768 for CLIP and PLIP, 512 for
CONCH, 1024 for UNI, 1536 for GigaPath, H-Optimus-0 and
MIDNIGHT, and 2560 for VIRCHOW2, whose feature is the class token
concatenated with the mean patch token. Where models support multiple variants we use the largest
publicly released checkpoint at the time of evaluation.

\section{Additional runtimes}
\label{app:runtimes}

Embedding extraction was performed on Google Colab NVIDIA A100 GPUs (80~GB
VRAM); the same figures are reported in the reproducibility checklist. Per-FM embedding cost across all six datasets ranges from approximately
0.6~GPU-hours (CLIP, the smallest model) to 3.4~GPU-hours (GigaPath and
H-Optimus-0, the largest). The total one-time embedding cost is approximately
15~GPU-hours. Once embeddings are cached, all 864 linear probe runs together
complete in under 30~minutes on a single CPU; bootstrap CI computation across
all 288 cells takes 26~seconds. Per-cell wall-clock for a single linear probe
run is dominated by the LBFGS convergence on the high-dimensional embedding
space and ranges from 0.3~seconds (CLIP on PCam at 1\% labels) to 12~seconds
(VIRCHOW2 on SPIDER-Breast at 100\% labels). The amortized cost structure is
the property that makes PRISM cheap to use as a comparison target for new
methods.

\section{Comparison with existing benchmarks}
\label{app:benchmarks}

\begin{table}[t]
\caption{Comparison of pathology FM benchmarks and adjacent calibration
studies. Only PRISM evaluates the joint interaction of FM choice, label
fraction, calibration, and cross-domain OOD shift on pathology FMs.
$^\dagger$~Image classification (not pathology); included as the closest
methodological precedent.}
\label{tab:related}
\centering
\small
\begin{tabular}{lccccc}
\toprule
Benchmark & FMs & Label fraction & Calibration & OOD & Year \\
\midrule
THUNDER~\cite{marza2026thunder}                              & 23     & few-shot only       & \checkmark         & adversarial only      & 2026 \\
HistoVL~\cite{majzoub2025histovl}                            & 4 VLMs & \ding{55}           & \checkmark         & \ding{55}             & 2025 \\
Neidlinger et al.~\cite{neidlinger2025}                      & 19     & \ding{55}           & \ding{55}          & \ding{55}             & 2025 \\
PathBench~\cite{ma2025pathbench}                             & 19     & \ding{55}           & \ding{55}          & \ding{55}             & 2025 \\
Lee et al.~\cite{lee2025adaptation}                          & 4      & few-shot only       & \ding{55}          & near/mid/out-domain   & 2025 \\
Nunes et al.~\cite{nunes2025beyond}                          & --     & \ding{55}           & \checkmark         & \ding{55}             & 2025 \\
Thiringer et al.~\cite{thiringer2026}                        & 14     & \ding{55}           & \checkmark         & scanner shift         & 2026 \\
Roschewitz et al.$^\dagger$~\cite{roschewitz2025calibration} & FT only & \ding{55}          & \checkmark         & \checkmark (natural)  & 2025 \\
\textbf{PRISM (ours)}                                        & \textbf{8} & \textbf{\checkmark (1--100\%)} & \textbf{\checkmark (+ recoverability)} & \textbf{\checkmark (4 + 20 pairs)} & \textbf{2026} \\
\bottomrule
\end{tabular}
\end{table}

A more detailed version of Table~\ref{tab:related} comparing PRISM to
THUNDER~\cite{marza2026thunder}, HistoVL~\cite{majzoub2025histovl},
Neidlinger et al.~\cite{neidlinger2025}, PathBench~\cite{ma2025pathbench},
Lee et al.~\cite{lee2025adaptation}, Nunes et al.~\cite{nunes2025beyond},
Thiringer et al.~\cite{thiringer2026},
Roschewitz et al.~\cite{roschewitz2025calibration},
W{\"o}lflein et al.~\cite{wolflein2023benchmarking},
{eva}~\cite{gatopoulos2024eva},
Patho-Bench~\cite{zhang2025pathobench}, and
Kang et al.~\cite{kang2023benchmarking} accompanies the code
release. The dimensions on which PRISM is
unique are: continuous label-fraction sweep (six points from 1\% to 100\%
versus discrete shot counts in THUNDER and Lee et al.); both raw and
post-hoc-recovered ECE (versus raw-only in THUNDER, HistoVL, and Thiringer);
and cross-domain transfer pairs (versus scanner-only in Thiringer, adversarial
in THUNDER, and none in Neidlinger / PathBench / HistoVL).

\section{Extended discussion on quantitative results}
\label{app:full_results}

\paragraph{What a label fraction consumes (six main datasets).} The fraction subsamples the
training split; the validation split used for temperature fitting is a
separate budget, and at 1\% labels it supplies 90--96\% of all labelled
examples a cell consumes. We therefore also report a \emph{proportional}
protocol in which validation is subsampled by the same fraction and seed:
AUROC and raw ECE are identical by construction, scaled ECE at 1\% labels
rises by 0.009, the decoupling contrast is essentially unchanged
($\Delta\rho = 0.69$ against 0.75), and the calibration ranking of FMs is
preserved (mean Kendall $\tau = 0.75$, no sign inversions). Because the
proportional budget leaves three validation samples on MHIST at 1\% labels,
scaled ECE at 1\% is read together with raw ECE, which uses no validation
labels (Section~\ref{sec:experiments}); absolute counts are in
Table~\ref{tab:label_budget}.

Per-(model, dataset, fraction) results for AUROC, ECE (raw and after
temperature scaling), Brier score, and macro-F1, reported as
mean~$\pm$~standard~deviation across the three seeds (42, 123, 456),
accompany the code release as one CSV file per (model, dataset). At label
fraction = 1.0, all three seeds train on identical data with the deterministic
LBFGS solver, yielding zero between-seed variance and confidence intervals
that collapse to point estimates by construction; we report this regime
transparently rather than masking it. That variance is concentrated where the label budget is smallest
and where the task is hardest: at 1\% labels the between-seed standard
deviation in AUROC is 0.055 on MHIST, 0.034 on LungHist700 and 0.026 on BRACS,
against 0.002 on CRC and 0.002 on PCam. It is also the larger source of
variation at that fraction, roughly twice the spread induced by sweeping the
regularization constant over two orders of magnitude, which is why we report it
alongside every low-fraction result rather than only the mean. It does not
account for the decoupling reported in Section~\ref{sec:experiments}: the
rank correlation between discrimination and calibration is negative at 1\%
labels within each of the three seeds taken separately ($-0.17$, $-0.25$,
$-0.10$), so the pattern is not produced by averaging across them.

\section{Implementation details}
\label{app:impl}

The \texttt{prism\_bench} package (\texttt{pip install prism-bench}) exposes a
\texttt{PRISMEvaluator} class plus a CLI (\texttt{prism evaluate},
\texttt{prism compare}). It operates on embeddings rather than models, so a
new FM is evaluated by extracting its features once with any pipeline and
running the protocol on CPU:

\begin{verbatim}
from prism_bench import PRISMEvaluator

ev = PRISMEvaluator(results_dir="results_v2")      # shipped reference results
res = ev.evaluate(X_train, y_train, X_test, y_test,
                  dataset="pcam", model_name="MyFM",
                  val_features=X_val, val_labels=y_val)  # for temperature
cmp = ev.compare(res, dataset="pcam", fraction=0.1) # ranked vs 8 reference FMs
\end{verbatim}

\texttt{evaluate} sweeps the six label fractions and three seeds by default
and returns one row per (fraction, seed) with AUROC, macro-F1, Brier, raw and
scaled ECE under both binning schemes, the fitted temperature and the
degeneracy share; without a validation split the temperature and scaled ECE
are left undefined rather than fitted on the test set. \texttt{compare} is
restricted to the six benchmark datasets, for which reference results exist.
The CLI mirrors this interface on \texttt{.npy} files. Random seeds
are fixed at \{42, 123, 456\}; \texttt{np.random.seed} is called immediately
before subset sampling to guarantee identical training subsets across FMs at
every (dataset, fraction, seed) cell. Package version pins, dataset download
instructions, and embedding generation scripts are included in the code
release.

\section{Impact of probe choice on calibration performance}
\label{app:probe}

All main-text results use $\ell_2$-regularized logistic regression as the
probe family. As Marza et al.~\cite{marza2026thunder} note, calibration is
probe-dependent, and we quantify that dependence rather than assume it away.
Repeating the full protocol with a multilayer perceptron of one hidden layer
of width 256, under identical splits, seeds and metric definitions, gives the
following. The MLP is uniformly better calibrated (mean ECE lower by 0.128 at
1\% labels and 0.039 at full supervision) and marginally better at
discrimination (mean AUROC higher by 0.004 to 0.015). Discriminative ordering
is largely preserved: the Kendall $\tau$ between the linear and MLP AUROC
rankings averages 0.85 across the 36 (dataset, fraction) cells. Calibration
ordering is not: $\tau$ averages 0.29 and inverts in sign in several cells,
reaching $-0.64$ on BRACS at 10\% labels. Of the qualitative findings, the
separation between discrimination and calibration persists in direction but
weakens in magnitude (pooled $\rho = 0.07$ at 1\% labels against $-0.24$
under the linear probe, and 0.42 against 0.52 at full supervision), and the
reverse OOD scaling of Section~\ref{sec:experiments} persists for all four
models it covers. One linear-probe pattern does \emph{not} persist: on
LungHist700 raw ECE \emph{rises} with label fraction under the linear probe
(UNI 0.213 to 0.313, GigaPath 0.186 to 0.312, H-Optimus-0 0.161 to 0.314),
whereas under the MLP it improves with additional labels for UNI ($0.144 \to 0.086$), GigaPath ($0.191 \to 0.067$) and
H-Optimus-0 ($0.114 \to 0.096$). That finding therefore belongs to the
linear-probe protocol and we state it as such. Taken with the regularization
sweep reported above, the pattern across both ablations is consistent:
in-distribution comparisons of discrimination between FMs are robust to probe
details while comparisons of calibration are not, which is the practical reason
a calibration-based index must fix its probe protocol before it can be used to
rank models.

\paragraph{LoRA check.} A small LoRA fine-tuning check~\cite{hu2022lora}
(UNI and H-Optimus-0 on MHIST and LungHist700, three fractions, three seeds;
36 runs) accompanies the code release. It is too small, and its training
budget too uneven across fractions, to support conclusions.

\paragraph{Regularization sweep in distribution.} On MHIST, LungHist700 and
BRACS we refit every probe at $C \in \{0.01, 0.1, 1, 10, 100\}$ (fractions
1\%, 10\% and 100\%, three seeds; raw ECE). Pooled within-dataset Spearman
correlations between AUROC rank and raw-ECE rank at 1\% and 100\% labels are
$-0.41$ and $-0.46$ at $C = 0.01$, $-0.08$ and $-0.54$ at $C = 0.1$, $-0.80$ and
$0.13$ at $C = 1$, $0.03$ and $0.10$ at $C = 10$, and $0.08$ and $0.45$ at
$C = 100$. The 1\%-versus-100\% contrast is therefore pronounced only at the
protocol value $C = 1$ and reverses under stronger regularization, which is
why Section~\ref{sec:protocols} states the decoupling as a property of the
fixed-probe pipeline. AUROC rankings are stable across the same sweep (mean
Kendall $\tau = 0.82$ against $C = 1$).

\paragraph{Regularization sweep under transfer.} We ran the regularization sweep in the transfer setting as well, over all four
pairs, all eight FMs, $C \in \{0.01, 0.1, 1, 10, 100\}$, three label fractions
and three seeds, for 1{,}440 further fits. Two results. First, the transfer
conclusions are insensitive to $C$: against $C = 1.0$ the mean Kendall $\tau$ is
0.71 for the OOD AUROC ranking with no sign inversions in 48 comparisons, and
0.59 for the OOD calibration ranking with two. The ranking instability reported
above is therefore an in-distribution phenomenon and does not carry into
transfer, although in \emph{magnitude} the asymmetry persists: the ratio of ECE
spread to AUROC spread across $C$ is above 1 on three of the four pairs, ranging
to 4.31 on PCam$\to$MHIST at 1\% labels. Across the four pairs, target raw
ECE rises from 1\% to 100\% source labels in 20, 27, 28, 27 and 22 of the 32
(model, pair) combinations at $C = 0.01, 0.1, 1, 10, 100$, while target
single-class collapse at 1\% labels falls from 82\% to 41\% of cells as $C$
grows, so the reverse-scaling direction is not produced by the protocol value
of $C$ alone.

\begin{figure}[t]
\centering
\includegraphics[width=\linewidth]{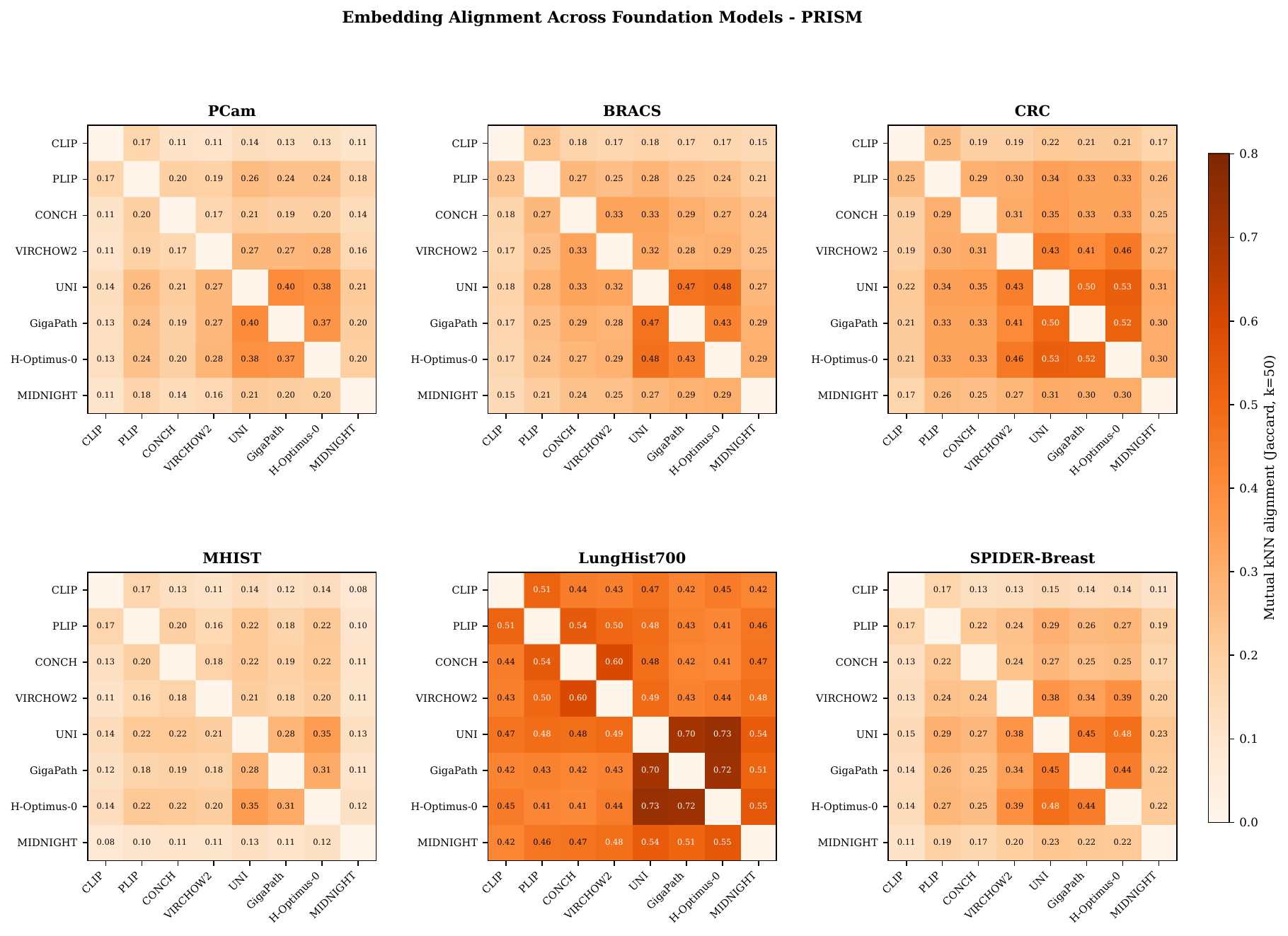}
\caption{\textbf{Embedding alignment across foundation models.} Mutual
$k$-nearest-neighbor alignment ($k=50$, Jaccard overlap of per-sample
neighbor sets) between every pair of FMs, computed on each test split.
Higher values indicate more similar embedding geometry. Alignment is
uniformly low on PCam, BRACS, and CRC, where every FM produces a
distinctive neighbor structure, and substantially higher on LungHist700
and SPIDER-Breast, where the four pathology SSL FMs (UNI,
GigaPath, H-Optimus-0, VIRCHOW2) form a tightly aligned subgroup. MHIST
sits in between, with moderate alignment that mirrors its position as
the binary task with the most cross-model disagreement. The diagonal is
omitted as self-alignment is identically 1.}
\label{fig:16_alignment_heatmap}
\end{figure}

\begin{figure}[t]
\centering
\includegraphics[width=0.75\linewidth]{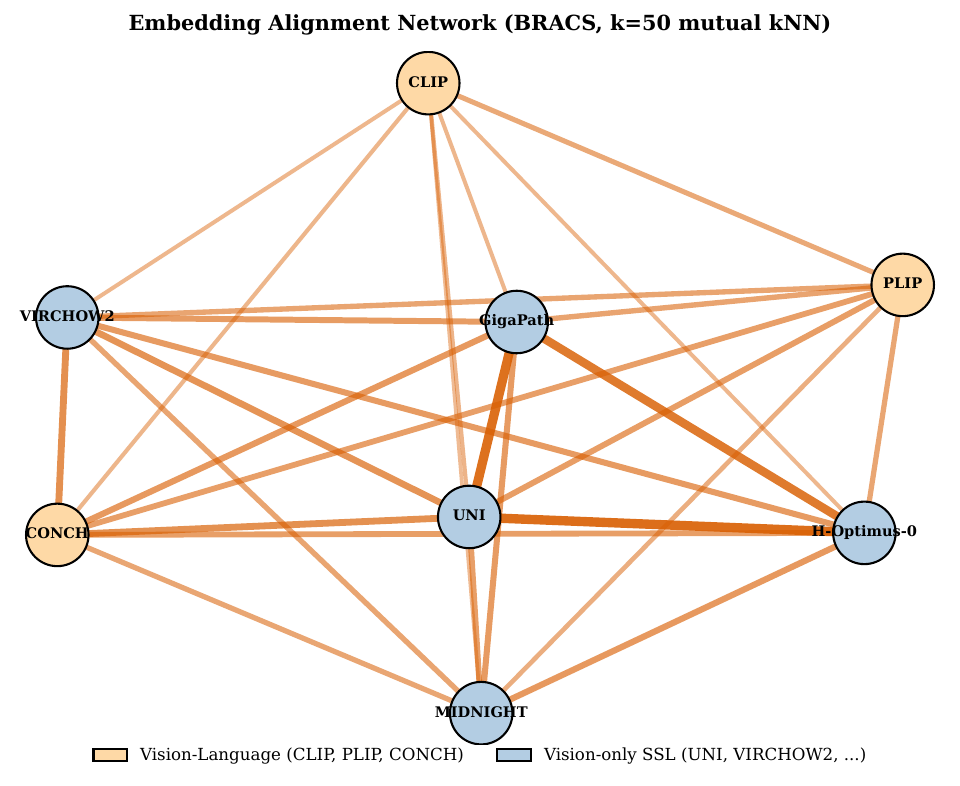}
\caption{\textbf{Embedding alignment network on BRACS.} Force-directed
layout of the eight FMs with edge thickness proportional to mutual
$k$-nearest-neighbor alignment ($k=50$). The strongest pairwise
alignment is between GigaPath and H-Optimus-0, with VIRCHOW2 and UNI
joining the same dense central cluster; CLIP and PLIP sit at the
periphery, consistent with their non-pathology and Twitter-pathology
pre-training corpora respectively. CONCH, despite being a vision-language
model, integrates into the central pathology-SSL cluster, indicating
that curated pathology image-caption pairs produce embedding geometry
closer to the vision-only SSL family than to CLIP or PLIP. The
complementary heatmaps for the remaining five datasets are reported in
Figure~\ref{fig:16_alignment_heatmap}.}
\label{fig:17_alignment_network}
\end{figure}

Second, and bearing on the mechanism proposed in
Section~\ref{sec:experiments}, the magnitude of the reverse OOD scaling effect
\emph{increases monotonically with} $C$. On MHIST$\to$PCam at 100\% source
labels, raw ECE rises from 0.221--0.246 across the four models at $C = 0.01$ to
0.443--0.498 at $C = 100$. Since $C$ is precisely the parameter that governs how
freely the probe may fit the source domain, this is quantitative support for the
over-specialization reading: constraining source-specific fitting suppresses the
effect in proportion. The effect is present at 19 of the 20 (model, $C$)
combinations; the exception is UNI at $C = 0.01$, a setting at which every FM
predicts a single class on the entire target test split and ECE is pinned at the
class prior at both ends of the label range, so no effect of any kind can be
expressed. Per-cell values accompany the code release. ECE is computed with fixed-width bins (15 bins) on the
test split. Because fixed-width binning can leave bins empty on small test
sets, we also report ECE under the equal-mass (adaptive) scheme of Nixon
et al.~\cite{nixon2019measuring}. The decoupling contrast is unchanged under
equal-mass bins (pooled $\rho = -0.24$ at 1\% labels and $0.56$ at full
supervision, against $-0.24$ and $0.52$ with equal-width bins), and the
paired values for every cell accompany the code release. For per-FM feature-space comparisons that
complement the calibration analysis we use mutual $k$-nearest-neighbour
alignment ($k=50$, Jaccard overlap of per-sample neighbour sets), reported in
Figures~\ref{fig:16_alignment_heatmap} and~\ref{fig:17_alignment_network};
centered kernel alignment~\cite{kornblith2019cka} is an alternative
probe-free measure that we do not compute here. The per-dataset alignment
matrices accompany the code release.

\section{Different pre-processing for SPIDER-Breast}
\label{app:spider}

SPIDER-Breast~\cite{nechaev2025spider} releases tiles at multiple
magnifications and sizes; we use the 224$\times$224 patches at 20$\times$
magnification to match the most common pathology FM training resolution.
The 18-class taxonomy is used as released, without merging or pruning.

\section{Camelyon17 hospital transfer: details}
\label{app:camelyon-details}

\paragraph{Data.} Camelyon17~\cite{bandi2018camelyon17} in the WILDS
release~\cite{koh2021wilds}, read from a parquet redistribution of the same
data and verified at 455{,}954 patches against the published release before
use. Five hospitals, one binary task, one label definition. We draw 20{,}000
patches per hospital, label-stratified at 50/50, and partition 14{,}000 /
3{,}000 / 3{,}000 into train, validation and test, stratified so that every
partition of every hospital is exactly balanced.

\paragraph{Why the split is at patch level.} Each hospital contributes about
ten slides and tumour patches concentrate in two or three of them. A
slide-disjoint partition therefore produced tumour prevalence between 0.1\% and
83\% across splits, ranging within a single hospital from 5.8\% in train to
81\% in test. Since ECE is prevalence-sensitive, cross-hospital calibration
comparisons under such a split would measure prevalence rather than
calibration. Patch-level stratification fixes prevalence at 50\% everywhere.
The cost is within-hospital slide overlap between train and test, which affects
only the in-distribution numbers; in the transfer grid, source and target are
different hospitals with disjoint patients and slides, so no leakage is
possible there. We report the in-distribution results on this dataset as an
upper bound.

\paragraph{Preprocessing.} Unlike the six main datasets, each FM here is
embedded under its own documented transform, resolved programmatically from the
model rather than imposed on it (Table~\ref{tab:cam-preproc}). Embeddings are
L2-normalised, matching the main protocol, so the regularisation constant means
the same thing on both datasets.

\begin{table}[h]
\caption{Preprocessing resolved from each model for the Camelyon17 extraction.}
\label{tab:cam-preproc}
\centering
\small
\begin{tabular}{lllr}
\toprule
Model & Resize & Normalization & Dim \\
\midrule
CLIP, PLIP & 224 & CLIP statistics & 768 \\
CONCH & model loader & model loader & 512 \\
UNI & 224 & ImageNet & 1024 \\
VIRCHOW2 & 224 & ImageNet & 2560 \\
GigaPath & 224 & ImageNet & 1536 \\
H-Optimus-0 & 256 & model-specific & 1536 \\
MIDNIGHT & 224 & (0.5, 0.5, 0.5) & 1536 \\
\bottomrule
\end{tabular}
\end{table}

\paragraph{Label budget.} At 1\% labels a cell trains on 140 patches. Under the
protocol used for the six main datasets the temperature is fitted on the full
3{,}000-patch validation split, so 95.5\% of the labels the cell consumes come
from validation; the share falls to 68\% at 10\% labels and 18\% at full
supervision. A proportional protocol, in which validation is subsampled by the
same fraction and seed, changes scaled ECE by at most 0.001 at any fraction and
leaves AUROC and raw ECE identical by construction. It is not the protocol we
report, however, because at 1\% labels it leaves 30 validation patches and the
fitted temperature reaches a search bound in 47\% of cells; the full-validation
temperature is stable at every fraction.

\begin{figure}[h]
\centering
\includegraphics[width=\linewidth]{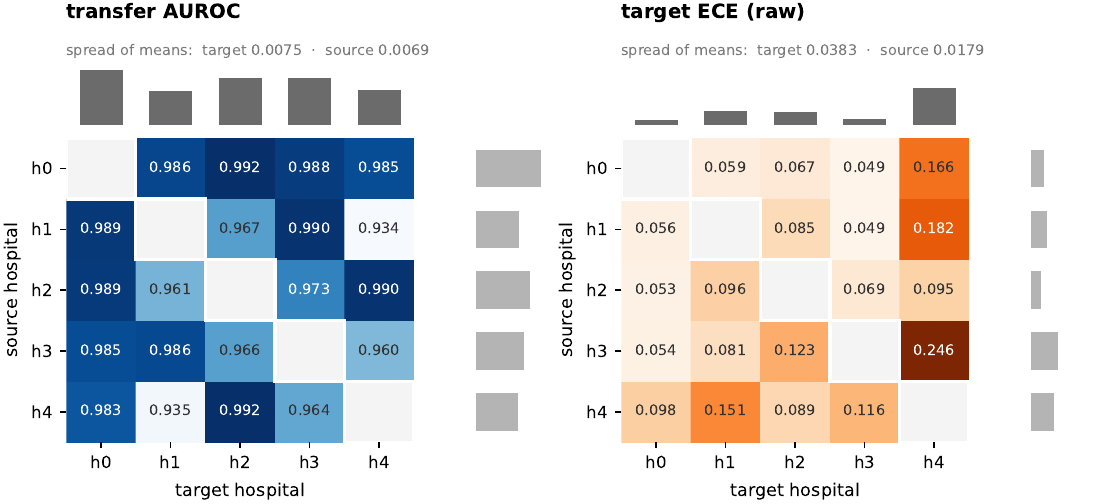}
\caption{\textbf{Calibration under transfer depends mostly on the target hospital.} Transfer
AUROC and raw target ECE at full supervision, averaged over the eight
foundation models, for each ordered pair of Camelyon17 hospitals. Marginal bars
give column and row means. The diagonal is empty because source and target are
always distinct. For raw ECE, column (target) means vary about twice as much
as row (source) means; for AUROC the two spreads are similar.}
\label{fig:hospital-matrix}
\end{figure}

\paragraph{Mechanism analysis.} The 192 (model, pair) combinations used in
Section~\ref{sec:experiments} combine the 32 from the four original pairs with
the 160 from the twenty hospital pairs. For each combination we fit the slope
of mean target ECE against $\log_{10}$ label fraction. Table~\ref{tab:cam-bands}
gives the rate at which that slope is positive, by transfer AUROC band.

\begin{table}[h]
\caption{Rate of rising target ECE by transfer AUROC, over all 192 (model,
pair) combinations from both transfer designs. The two middle bands contain
combinations from both designs.}
\label{tab:cam-bands}
\centering
\small
\begin{tabular}{lrrrl}
\toprule
Transfer AUROC & $n$ & Rising & Rate & Mean $\Delta$ECE \\
\midrule
$[0.00, 0.55)$ & 16 & 14 & 88\% & $+0.081$ \\
$[0.55, 0.70)$ & 17 & 15 & 88\% & $+0.089$ \\
$[0.70, 0.85)$ & 7 & 6 & 86\% & $+0.110$ \\
$[0.85, 0.95)$ & 13 & 1 & 8\% & $-0.074$ \\
$[0.95, 1.00]$ & 139 & 2 & 1\% & $-0.172$ \\
\bottomrule
\end{tabular}
\end{table}

\paragraph{Per-model transfer on Camelyon17.} Averaged over the twenty pairs
and the six label fractions, OOD AUROC and the number of pairs on which target ECE rises: CLIP 0.849 and 8;
PLIP 0.966 and 1; CONCH 0.984 and 0; MIDNIGHT 0.989 and 0; GigaPath 0.992 and
1; H-Optimus-0 0.992 and 0; UNI 0.994 and 0; VIRCHOW2 0.995 and 0. The Spearman
correlation between the two columns is $-0.91$ ($p = 0.002$).

\paragraph{Availability.} Split indices, the balanced re-split index map, the
preprocessing table, per-cell results for both grids and the mechanism summary
accompany the code release.

\section{Additional experimental results}
\label{app:additional}

\begin{figure}[t]
\centering
\includegraphics[width=0.95\linewidth]{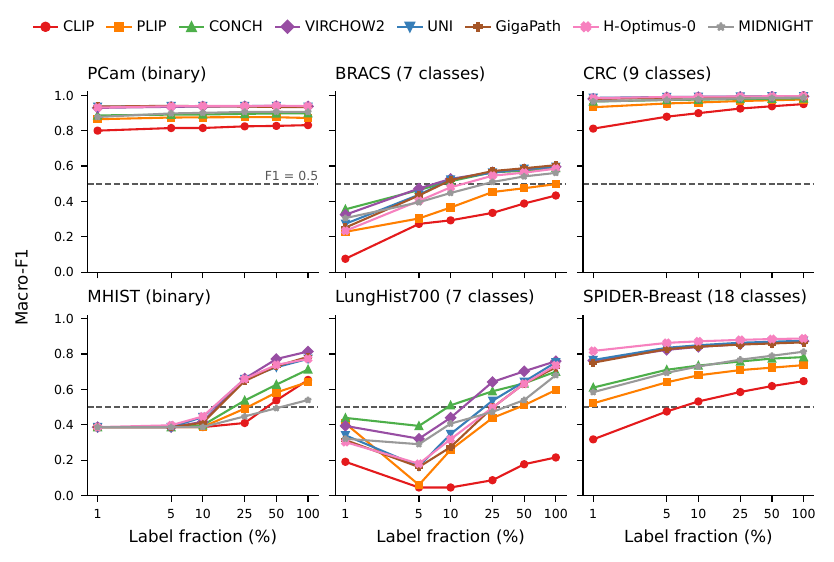}
\caption{\textbf{Breaking points.} Macro-F1 versus label fraction for each
(model, dataset) pair. PCam and CRC are close to saturation (macro-F1 at
least 0.80 already at 1\% labels); \{BRACS, MHIST,
LungHist700\} expose label-efficiency differences across FM classes. CLIP and
PLIP never break F1 = 0.5 on BRACS regardless of label budget.}
\label{fig:4_breaking_point}
\end{figure}

\begin{figure}[t]
\centering
\includegraphics[width=0.95\linewidth]{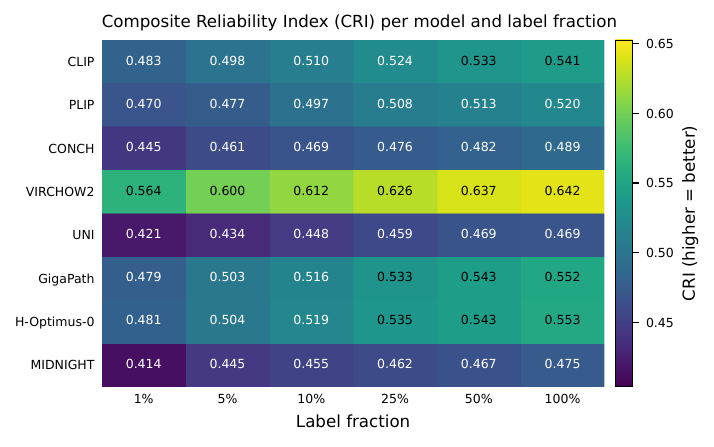}
\caption{\textbf{Composite Reliability Index.} CRI per (model, label fraction)
cell, aggregating discrimination, calibration and OOD stability with
$\mathrm{OOD\_Stability}$ as defined in Equation~\ref{eq:cri}. VIRCHOW2 leads
in both regimes (0.56 at 1\% labels, 0.64 at 100\%). Two features of the
index are worth reading off the figure rather than inferring from it. The
spread between the best and worst model is 0.15 at 1\% labels and 0.17 at
full supervision, so the index separates models less sharply than AUROC alone
does. And CLIP places second at 1\% labels despite being last on AUROC on
five of six datasets, because a ratio-based stability term rewards models that
had little in-distribution performance to lose under transfer. We read CRI as
a screening heuristic subject to that caveat, not as a ranking.}
\label{fig:6_cri}
\end{figure}

\begin{figure}[t]
\centering
\includegraphics[width=0.95\linewidth]{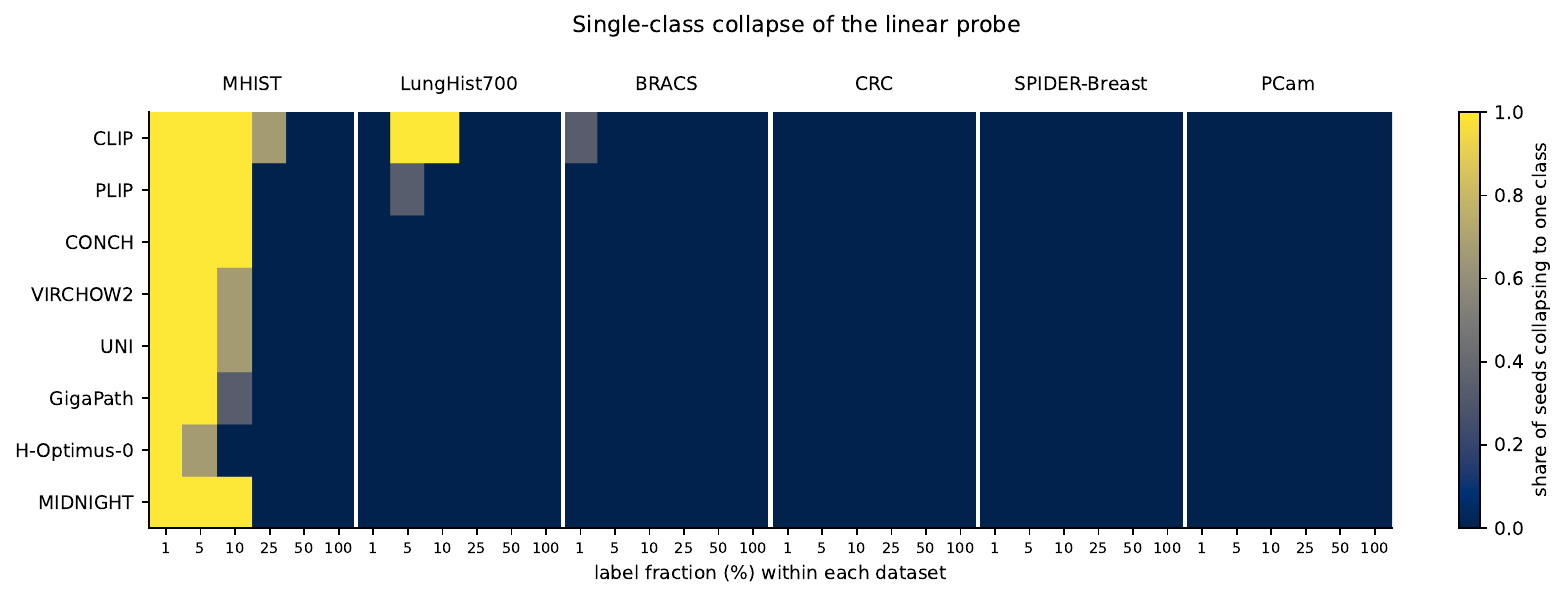}
\caption{\textbf{Single-class collapse of the linear probe.} Share of the
three seeds in which the probe assigns more than 99\% of the test set to one
class, per (model, dataset, label fraction). On MHIST every FM collapses at
1\% labels and most remain collapsed at 5\%; on PCam and CRC collapse never
occurs. Stratified sampling does not remove the effect, so it is not purely an
artefact of subset selection: at these fractions the probe has enough examples
of each class and still does not move its decision threshold off the majority.
Where a cell is dark, AUROC may remain informative while macro-F1 and ECE
describe a classifier that makes one prediction.}
\label{fig:18_degeneracy}
\end{figure}

\begin{figure}[t]
\centering
\includegraphics[width=\linewidth]{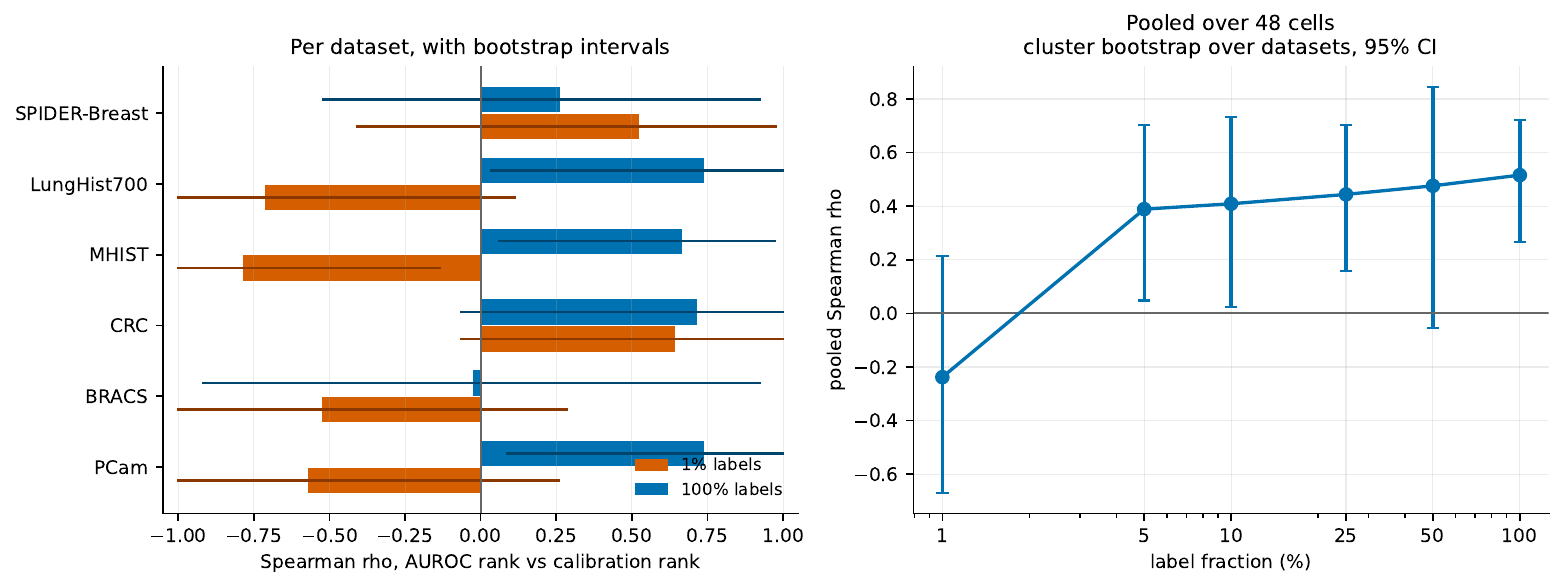}
\caption{\textbf{Rank correlation between discrimination and calibration.}
Left: per-dataset Spearman $\rho$ between AUROC rank and calibration rank at
1\% and 100\% labels, with bootstrap intervals over the eight models. Right:
the same quantity pooled over all 48 (model, dataset) cells at each label
fraction, with intervals from a cluster bootstrap that resamples whole
datasets rather than models, since models within a dataset are not
independent. The pooled estimate rises from $-0.24$ at 1\% labels to $0.52$
at full supervision; the contrast between the two regimes is significant
($\Delta\rho = 0.62$, 95\% CI $[0.08, 1.13]$ when the bootstrap resamples seeds
as well as datasets) even though the individual per-fraction estimates do not
all exclude zero on their own.}
\label{fig:19_rank_correlation}
\end{figure}

This appendix presents the supporting figures that complement the
main-text figures of Section~\ref{sec:experiments}.
Figure~\ref{fig:18_degeneracy} reports the collapse audit that qualifies the
breaking-point analysis, and Figure~\ref{fig:19_rank_correlation} gives the
rank-correlation statistics underlying the decoupling claim.
Figure~\ref{fig:4_breaking_point} gives the breaking-point curves and
Figure~\ref{fig:6_cri} the Composite Reliability Index heatmap discussed in
Section~\ref{sec:experiments}.
Table~\ref{tab:oodmap} gives the binary label mapping of the four original transfer pairs.

\begin{table}[t]
\caption{Binary label mapping for the four transfer pairs. The positive class
is defined differently on each side, so these settings combine covariate shift
with a change of label definition.}
\label{tab:oodmap}
\centering
\small
\begin{tabular}{lll}
\toprule
Pair & Source positive & Target positive \\
\midrule
PCam$\to$MHIST  & tumour tissue, lymph node & SSA polyp \\
MHIST$\to$PCam  & SSA polyp                 & tumour tissue, lymph node \\
CRC$\to$BRACS   & TUM                       & IC, DCIS \\
BRACS$\to$CRC   & IC, DCIS                  & TUM \\
\bottomrule
\end{tabular}
\end{table}

\begin{figure}[t]
\centering
\includegraphics[width=0.95\linewidth]{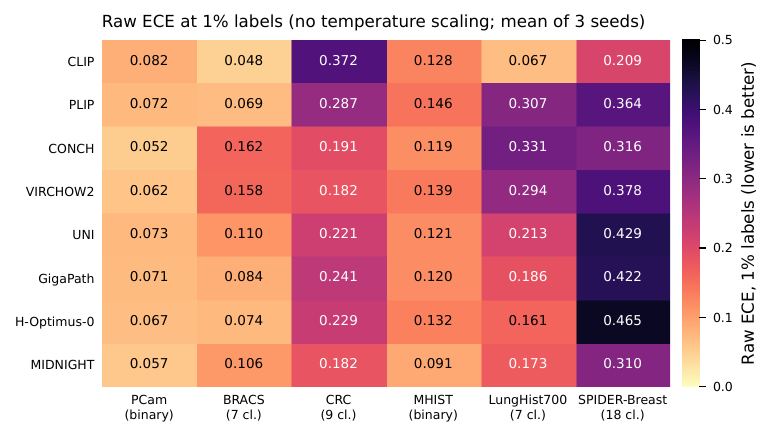}
\caption{\textbf{Raw miscalibration at 1\% labels.} Raw ECE for every
(model, dataset) cell. On SPIDER-Breast raw ECE reaches 0.42--0.47 for UNI,
GigaPath and H-Optimus-0 against 0.21 for CLIP. The fitted temperatures in
these cells are below 1 (Figure~\ref{fig:14_scaling_law}), so this is
under-confidence of the regularized linear probe, not over-confidence, and
temperature scaling largely removes it (Figure~\ref{fig:5_recoverability}).
Complementary to Figure~\ref{fig:7_decoupling}: AUROC rankings hide the
magnitude of raw miscalibration.}
\label{fig:11_raw_miscalibration}
\end{figure}

\begin{figure}[t]
\centering
\includegraphics[width=0.95\linewidth]{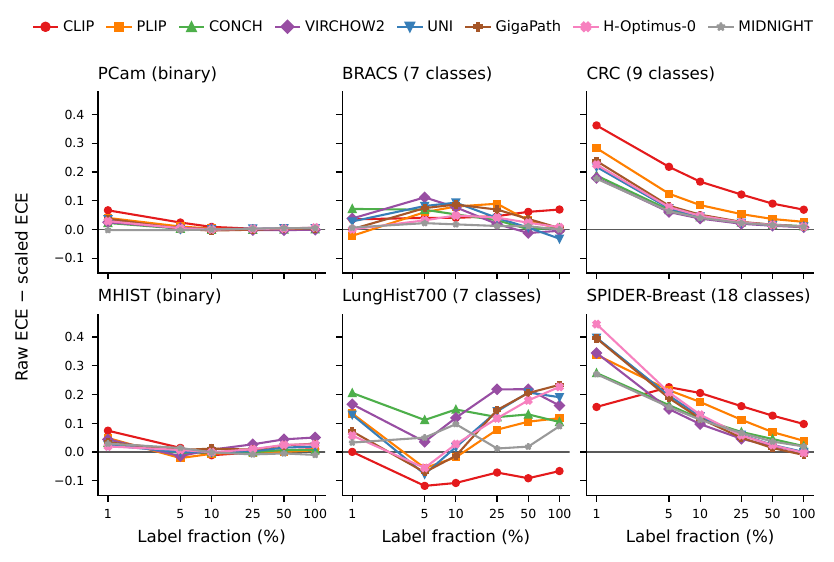}
\caption{\textbf{Calibration recoverability via temperature scaling.}
Improvement in ECE (raw minus scaled) per (model, dataset) cell, plotted
against label fraction. Recoverability ranges from 0.45 (H-Optimus-0
on SPIDER-Breast at 1\% labels, raw ECE 0.465 to scaled 0.020) to zero or
slightly negative (UNI on MHIST at 5\% and 10\% labels); scaling fails to
reduce ECE in 49 of the 288 cells. Practitioners can use this map to anticipate
when a single scalar $T$ suffices and when post-hoc calibration is inert.}
\label{fig:5_recoverability}
\end{figure}

\begin{figure}[t]
\centering
\includegraphics[width=0.95\linewidth]{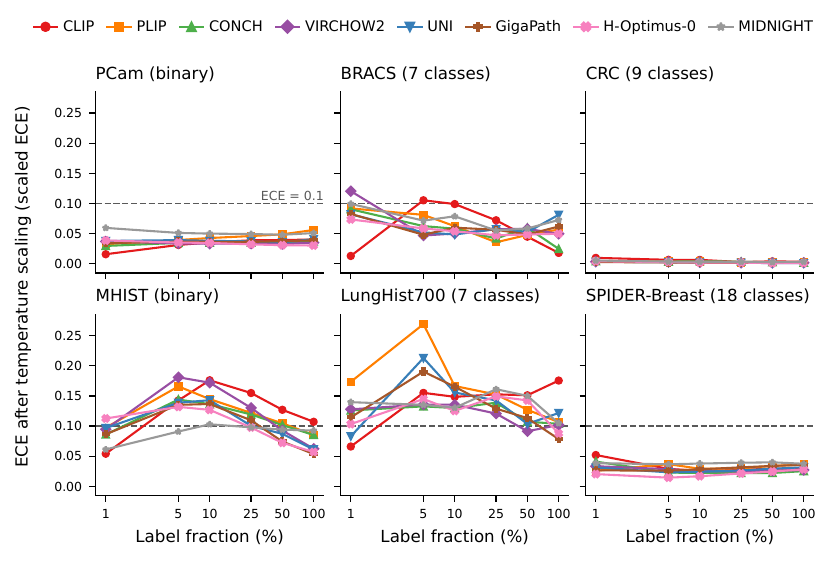}
\caption{\textbf{ECE after temperature scaling across label fractions.} ECE
per dataset as a function of label fraction, after applying the per-cell
temperature $T$ fitted on the validation split. From 5\% labels upward,
scaling keeps ECE below 0.1 for every FM on PCam, CRC and SPIDER-Breast and
in all but one BRACS cell, but not on MHIST and LungHist700, where 62 of the
80 cells remain above 0.1. Scaling also does \emph{not} repair the OOD
calibration failure documented in Figure~\ref{fig:10_ood_ece}.}
\label{fig:12_scaled_ece}
\end{figure}

\begin{figure}[t]
\centering
\includegraphics[width=0.95\linewidth]{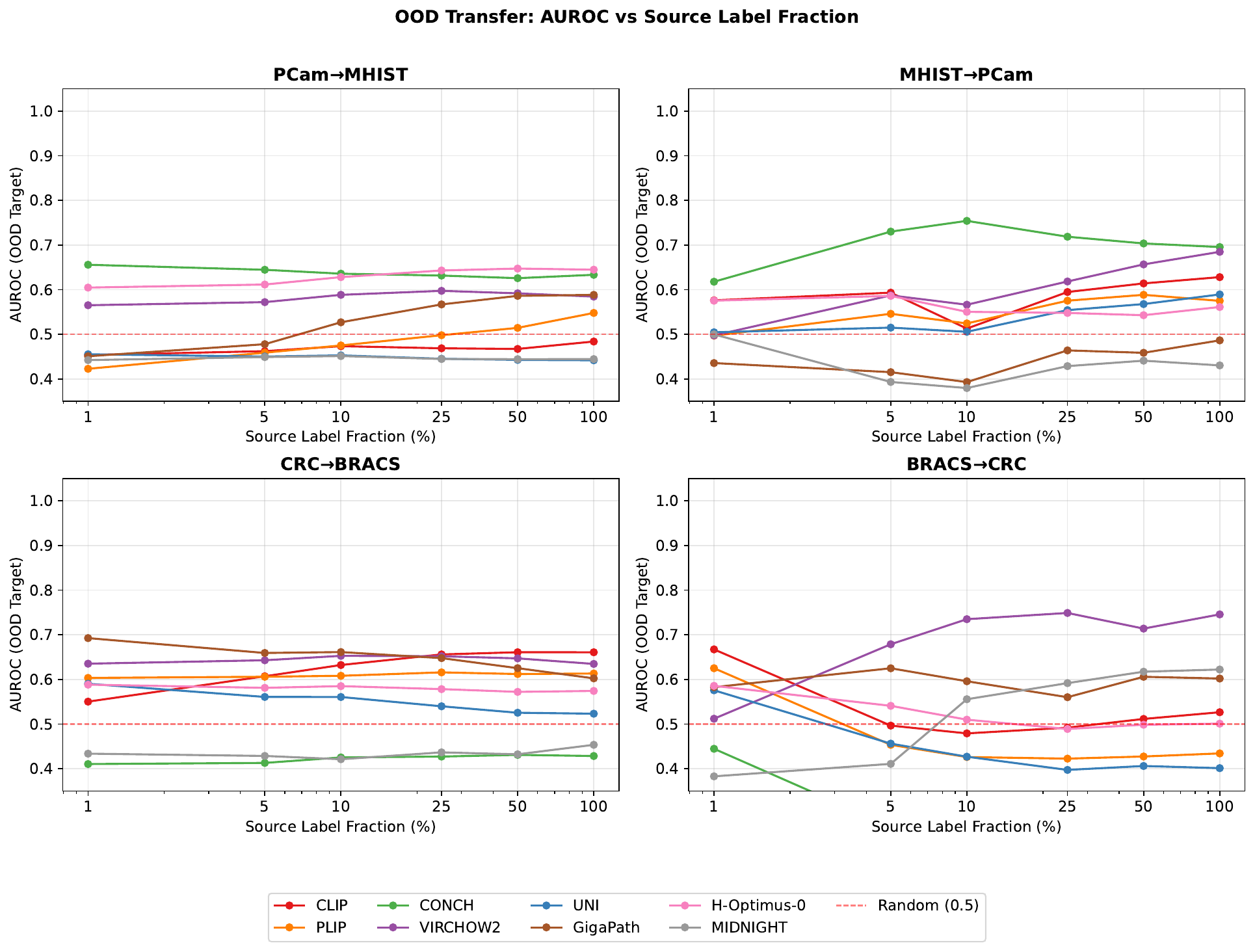}
\caption{\textbf{OOD transfer: AUROC versus source label fraction.} Companion
to Figure~\ref{fig:10_ood_ece}, showing OOD discrimination on the four transfer
pairs. On MHIST$\to$PCam, several FMs hold AUROC near random ($0.4$--$0.5$)
across the entire 1--100\% range while their OOD ECE rises monotonically:
confidence increasing without competence.}
\label{fig:8_ood_auroc_curves}
\end{figure}

\begin{figure}[t]
\centering
\includegraphics[width=0.95\linewidth]{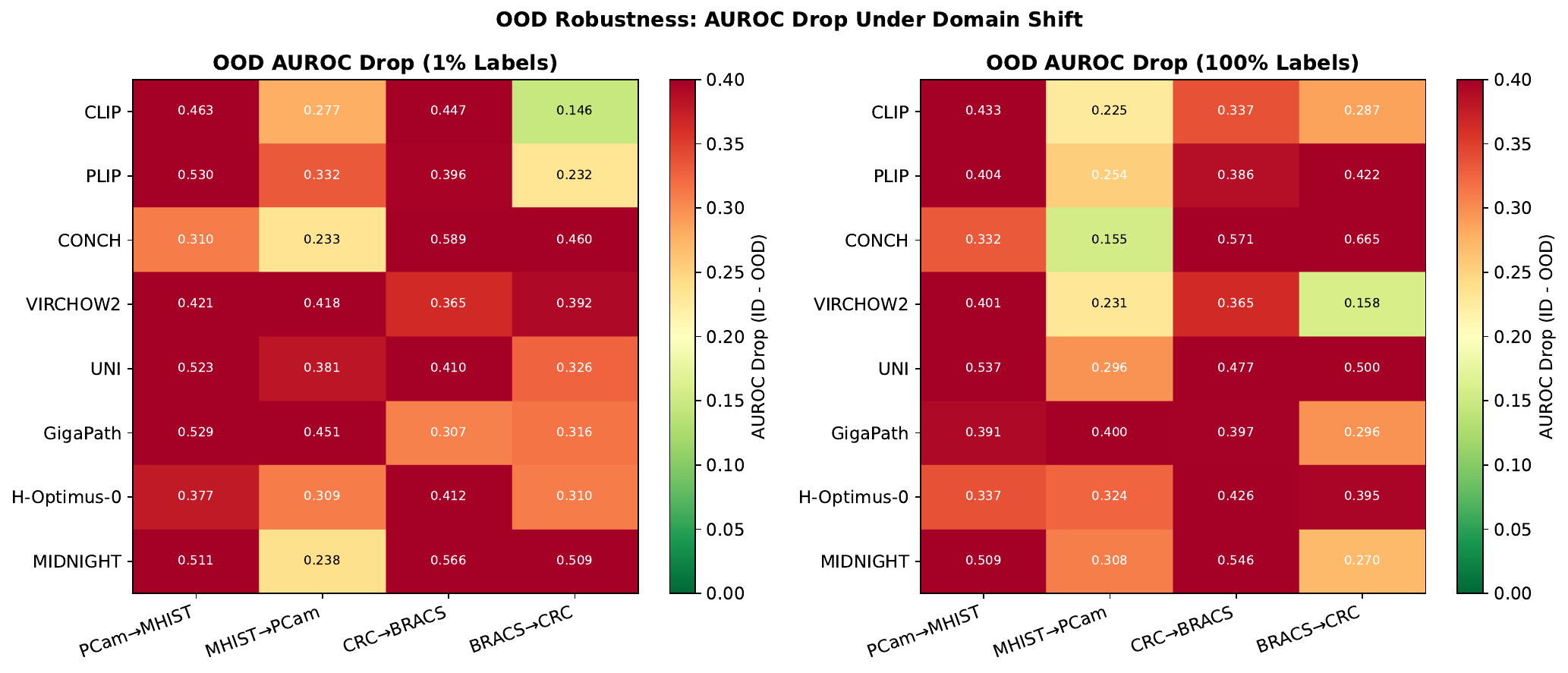}
\caption{\textbf{OOD AUROC drop heatmap.} Per (model, transfer pair) drop
in AUROC from in-distribution to OOD, at 1\% and 100\% labels. Drops above
0.4 (e.g., CONCH on CRC$\to$BRACS at 100\% labels) indicate that the
model's discrimination collapses near random under domain shift, even before
calibration is considered.}
\label{fig:9_ood_auroc_drop}
\end{figure}

\begin{figure}[t]
\centering
\includegraphics[width=0.95\linewidth]{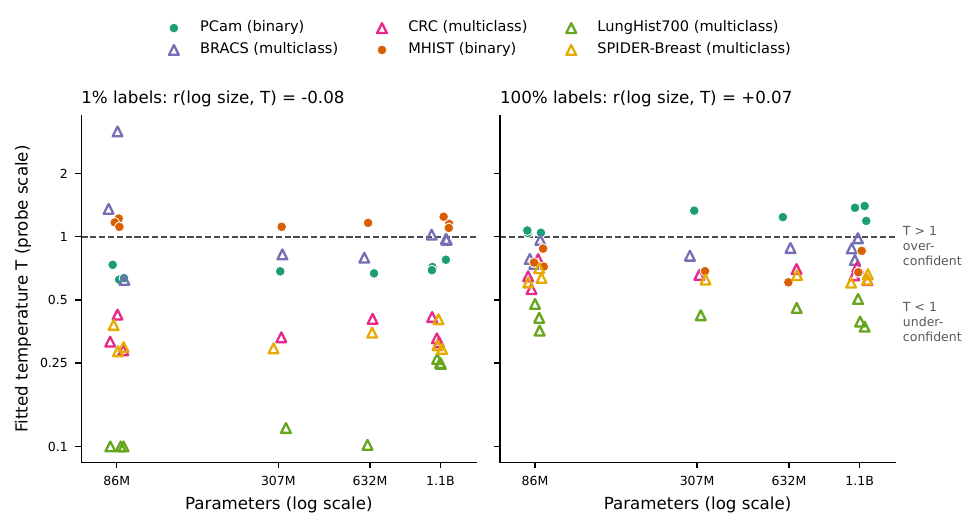}
\caption{\textbf{Fitted temperature versus model size.} Validation-fitted
temperature $T$ (on the probe's own logit scale) against parameter count, at
1\% and 100\% labels; $T > 1$ indicates an over-confident probe and $T < 1$
an under-confident one. The
correlation between $\log$ size and $T$ is negligible ($r = -0.08$ at 1\%
labels, $r = +0.07$ at 100\%). Task type dominates: $T < 1$ in 189 of
192 multiclass cells, whereas binary cells straddle $T = 1$. On LungHist700 at
1\% labels $T$ reaches the lower search bound (0.1) for four FMs, so scaled ECE
there is bound-limited. This pattern mirrors the
absence of a clean ``bigger is better'' rule in CRI rankings
(Figure~\ref{fig:6_cri}).}
\label{fig:14_scaling_law}
\end{figure}

\begin{figure}[t]
\centering
\includegraphics[width=0.95\linewidth]{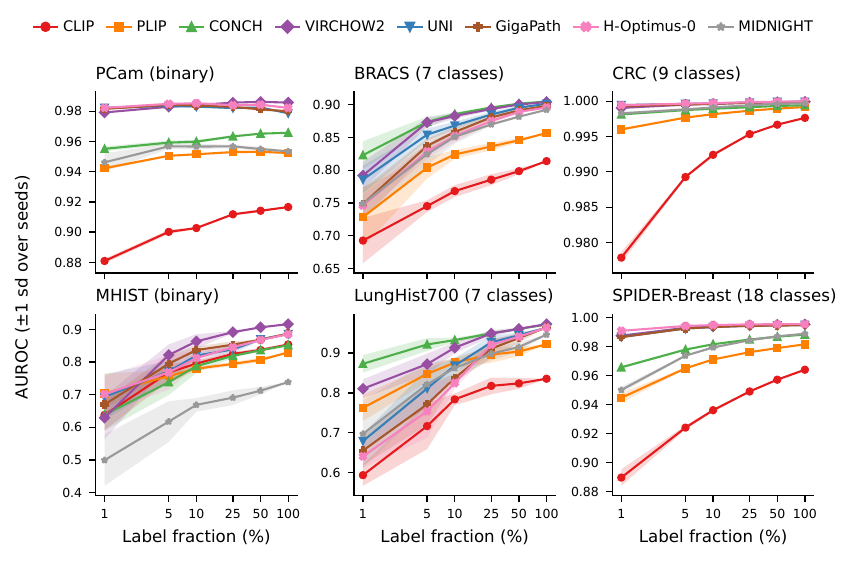}
\caption{\textbf{AUROC versus label fraction.} Companion to
Figure~\ref{fig:4_breaking_point}, showing AUROC instead of macro-F1 across the
same six datasets. Top-4 pathology SSL models (VIRCHOW2, UNI, GigaPath,
H-Optimus-0) form a tight cluster from 1\% labels upward, while CLIP, PLIP,
and MIDNIGHT separate visibly on the harder triplet
\{BRACS, MHIST, LungHist700\}. Shaded bands show $\pm 1$ standard deviation
across the three random seeds.}
\label{fig:8_auroc_curves}
\end{figure}

\begin{figure}[t]
\centering
\includegraphics[width=0.95\linewidth]{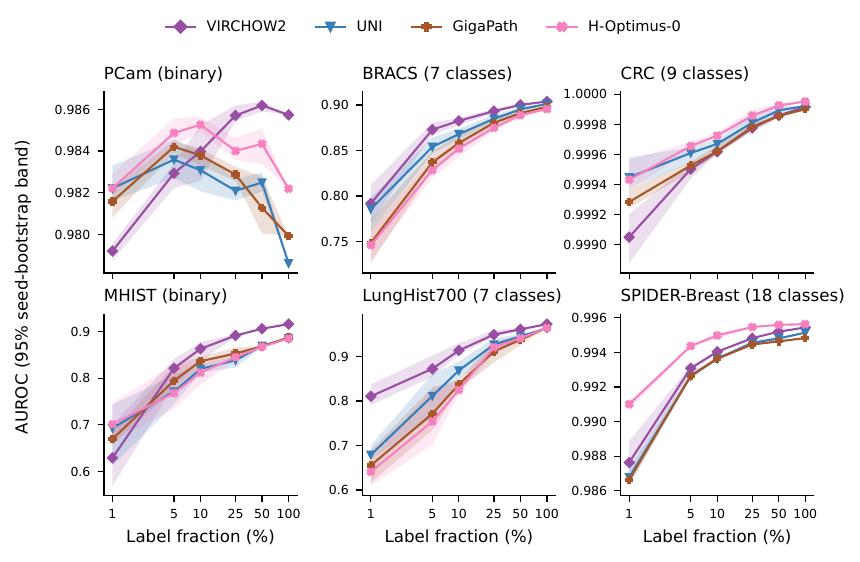}
\caption{\textbf{Bootstrap 95\% confidence intervals.} AUROC for the four
strongest pathology SSL FMs (VIRCHOW2, UNI, GigaPath, H-Optimus-0) across all
six datasets, with 1000-resample bootstrap CI bands computed over the three
random seeds. With only three seeds these bands describe seed variability, not
sampling uncertainty of the test set. Band width collapses at 100\% labels
because the deterministic LBFGS solver yields zero between-seed variance, as
noted in Appendix~\ref{app:full_results}.}
\label{fig:9_bootstrap_ci}
\end{figure}

\begin{figure}[t]
\centering
\includegraphics[width=0.95\linewidth]{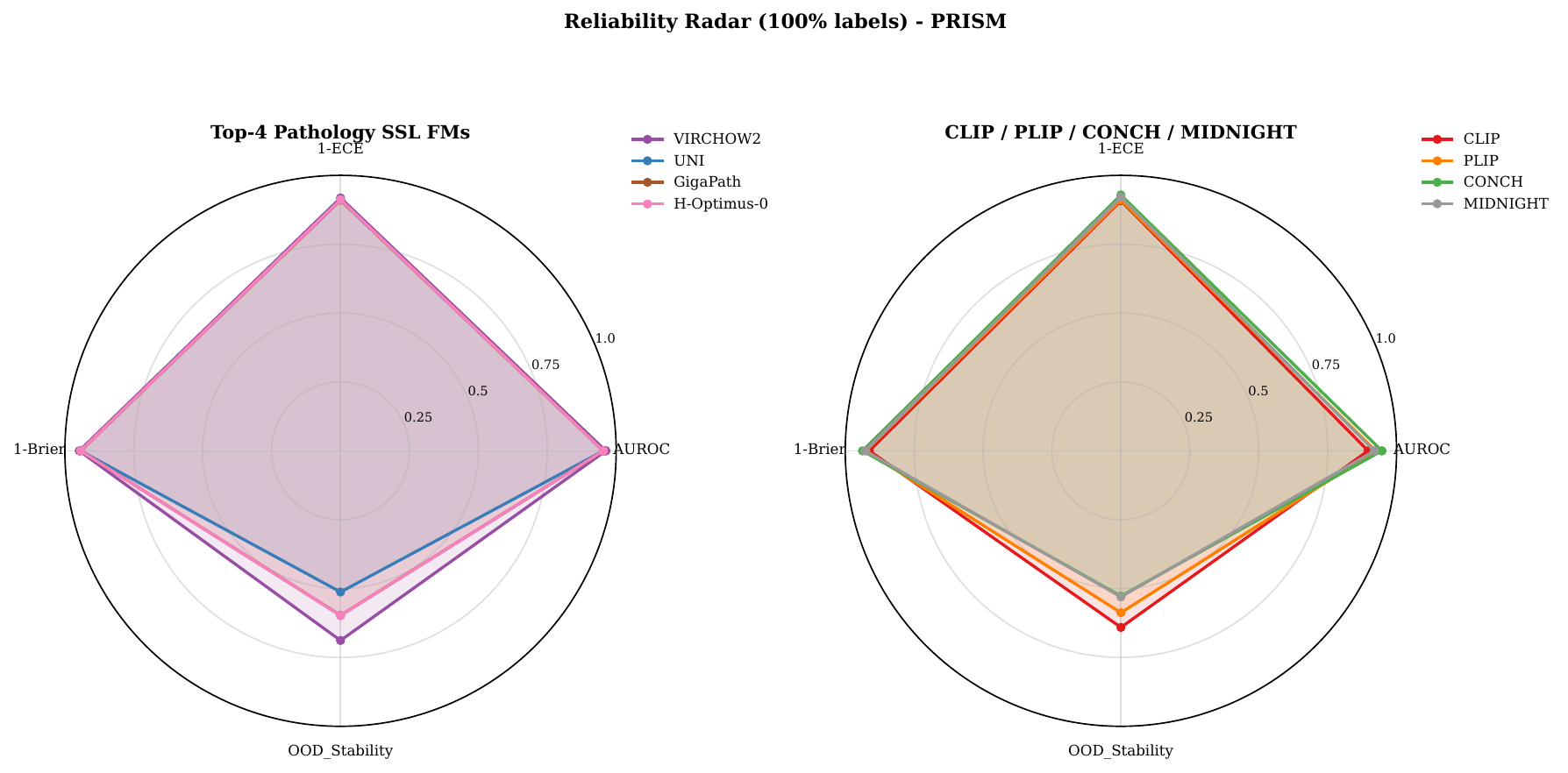}
\caption{\textbf{Reliability radar.} Per-model summary at 100\% labels
across the four reliability axes: AUROC (in-distribution, mean over 6
datasets), $1-\mathrm{ECE}_{\mathrm{raw}}$, $1-\mathrm{Brier}$, and
OOD\_Stability (mean OOD AUROC divided by mean in-distribution AUROC across
the four transfer pairs, clipped to $[0,1]$). The top-4 panel shows the
four large pathology SSL FMs; the bottom-4 panel shows
CLIP, PLIP, CONCH, and MIDNIGHT for visual separation.}
\label{fig:13_radar}
\end{figure}

\begin{figure}[t]
\centering
\includegraphics[width=0.85\linewidth]{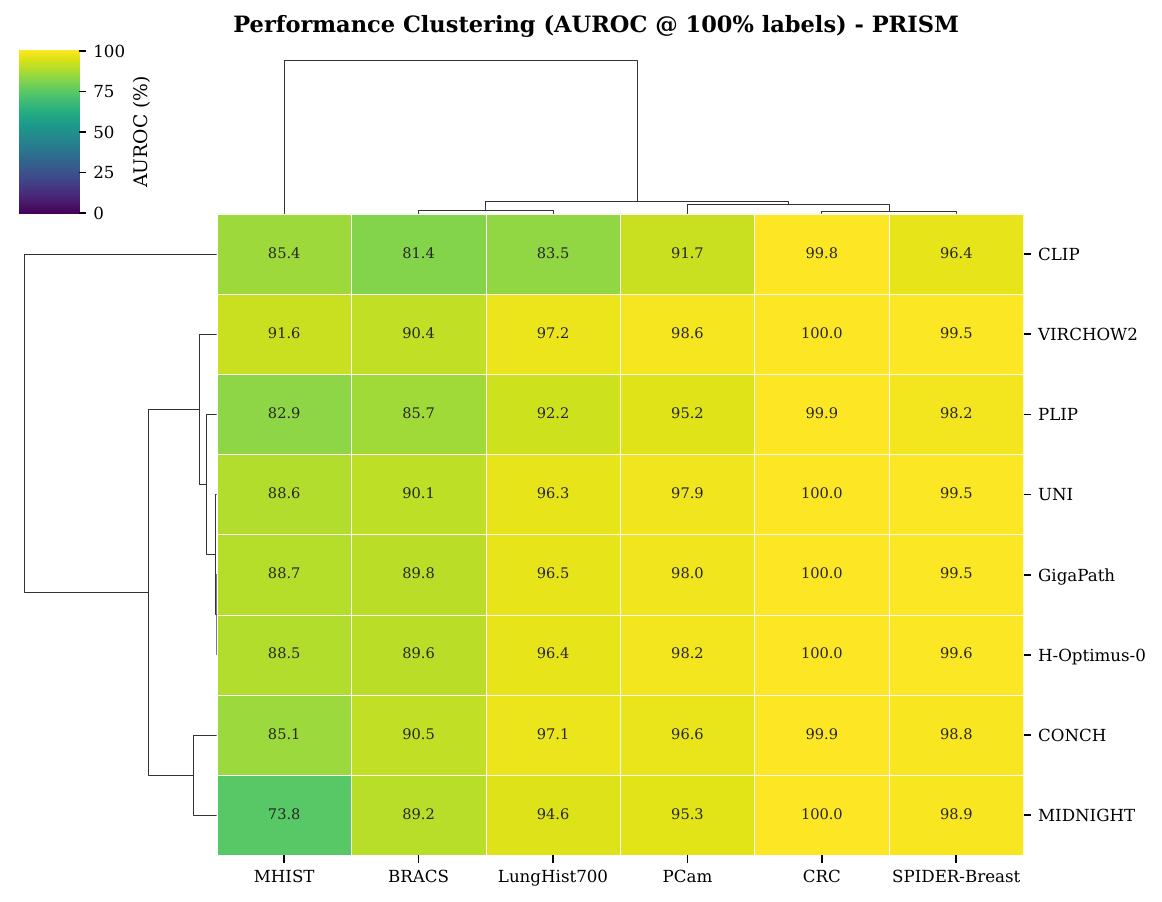}
\caption{\textbf{Performance clustering.} Hierarchical clustering of the
$8 \times 6$ AUROC matrix at 100\% labels using
$1-\rho$ correlation distance with average linkage on both axes. The
dendrogram on the rows separates the four pathology SSL FMs (VIRCHOW2, UNI,
GigaPath, H-Optimus-0) from CLIP and the vision-language pathology models
(PLIP, CONCH), with MIDNIGHT joining the lighter-pretraining cluster
despite its 1.1B parameter count. The grouping confirms that pre-training
corpus rather than parameter count is the dominant predictor of
in-distribution AUROC ranking on PRISM datasets.}
\label{fig:15_perf_clustering}
\end{figure}

\section{Detailed per-(model, dataset) results}
\label{app:detailed_tables}
This appendix presents the detailed per-(model, dataset) numerical results that underlie the heatmaps and curves in the main paper. All values are mean $\pm$ standard deviation across the three random seeds (42, 123, 456). At label fraction = 1.0 the deterministic LBFGS solver yields zero between-seed variance, which is reported as a point estimate.
\begin{table}[!htbp]
\centering
\small
\caption{AUROC at 100\% label fraction (mean $\pm$ std over 3 seeds). Best per dataset in bold.}
\label{tab:auroc_100}
\begin{tabular}{lcccccc}
\toprule
\textbf{Model} & \textbf{PCam} & \textbf{BRACS} & \textbf{CRC} & \textbf{MHIST} & \textbf{LungHist700} & \textbf{SPIDER-Breast} \\
\midrule
CLIP & 0.917\,$\pm$\,0.000 & 0.814 & 0.998\,$\pm$\,0.000 & 0.854 & 0.835 & 0.964\,$\pm$\,0.000 \\
PLIP & 0.952 & 0.857 & 0.999 & 0.829 & 0.922 & 0.982\,$\pm$\,0.000 \\
CONCH & 0.966\,$\pm$\,0.000 & \textbf{0.905} & 0.999 & 0.851 & 0.971 & 0.988\,$\pm$\,0.000 \\
VIRCHOW2 & \textbf{0.986} & 0.904 & 1.000 & \textbf{0.916} & \textbf{0.972} & 0.995 \\
UNI & 0.979\,$\pm$\,0.000 & 0.901 & 1.000 & 0.886 & 0.963 & 0.995 \\
GigaPath & 0.980 & 0.898 & 1.000 & 0.887 & 0.965 & 0.995\,$\pm$\,0.000 \\
H-Optimus-0 & 0.982 & 0.896 & \textbf{1.000} & 0.885 & 0.964 & \textbf{0.996} \\
MIDNIGHT & 0.953 & 0.892 & 1.000 & 0.738 & 0.946 & 0.989\,$\pm$\,0.000 \\
\bottomrule
\end{tabular}
\end{table}

At full label availability, in-distribution AUROC is largely saturated for the
pathology-pretrained vision SSL models, with VIRCHOW2, UNI, GigaPath, and
H-Optimus-0 all reaching 1.000 on CRC and clustering above 0.98 on PCam and
SPIDER-Breast. The two task families that retain headroom are MHIST (binary
HP/SSA) and BRACS (7-class breast carcinoma subtyping), where VIRCHOW2 attains
the top AUROC on MHIST (0.916) and CONCH leads on BRACS (0.905). MIDNIGHT,
trained on roughly an order of magnitude less data than VIRCHOW2, achieves
parity on PCam, CRC, and SPIDER-Breast but drops to 0.738 on MHIST, the lowest
score in the column. The compact spread on the saturated datasets is the
property that motivates the more discriminative protocols of
Section~\ref{sec:protocols}: at 100\% labels, raw AUROC alone provides limited
signal for ranking pathology FMs.

\begin{table}[!htbp]
\centering
\small
\caption{AUROC at 1\% label fraction (mean $\pm$ std over 3 seeds). Best per dataset in bold.}
\label{tab:auroc_1}
\begin{tabular}{lcccccc}
\toprule
\textbf{Model} & \textbf{PCam} & \textbf{BRACS} & \textbf{CRC} & \textbf{MHIST} & \textbf{LungHist700} & \textbf{SPIDER-Breast} \\
\midrule
CLIP & 0.881\,$\pm$\,0.001 & 0.693\,$\pm$\,0.035 & 0.978\,$\pm$\,0.001 & 0.638\,$\pm$\,0.047 & 0.594\,$\pm$\,0.027 & 0.890\,$\pm$\,0.006 \\
PLIP & 0.942\,$\pm$\,0.001 & 0.728\,$\pm$\,0.046 & 0.996\,$\pm$\,0.000 & \textbf{0.704\,$\pm$\,0.061} & 0.762\,$\pm$\,0.033 & 0.945\,$\pm$\,0.003 \\
CONCH & 0.955\,$\pm$\,0.002 & \textbf{0.823\,$\pm$\,0.021} & 0.998\,$\pm$\,0.000 & 0.638\,$\pm$\,0.048 & \textbf{0.873\,$\pm$\,0.022} & 0.966\,$\pm$\,0.001 \\
VIRCHOW2 & 0.979\,$\pm$\,0.000 & 0.792\,$\pm$\,0.021 & 0.999\,$\pm$\,0.000 & 0.629\,$\pm$\,0.066 & 0.810\,$\pm$\,0.024 & 0.988\,$\pm$\,0.001 \\
UNI & 0.982\,$\pm$\,0.001 & 0.786\,$\pm$\,0.021 & \textbf{0.999\,$\pm$\,0.000} & 0.693\,$\pm$\,0.069 & 0.679\,$\pm$\,0.018 & 0.987\,$\pm$\,0.000 \\
GigaPath & 0.982\,$\pm$\,0.001 & 0.748\,$\pm$\,0.024 & 0.999\,$\pm$\,0.000 & 0.670\,$\pm$\,0.013 & 0.655\,$\pm$\,0.037 & 0.987\,$\pm$\,0.000 \\
H-Optimus-0 & \textbf{0.982\,$\pm$\,0.001} & 0.746\,$\pm$\,0.016 & 0.999\,$\pm$\,0.000 & 0.701\,$\pm$\,0.055 & 0.641\,$\pm$\,0.029 & \textbf{0.991\,$\pm$\,0.000} \\
MIDNIGHT & 0.946\,$\pm$\,0.006 & 0.749\,$\pm$\,0.022 & 0.998\,$\pm$\,0.000 & 0.499\,$\pm$\,0.079 & 0.697\,$\pm$\,0.084 & 0.950\,$\pm$\,0.002 \\
\bottomrule
\end{tabular}
\end{table}

At 1\% labels the ranking restructures considerably. CONCH, the curated
pathology vision-language model, leads on BRACS (0.823) and LungHist700
(0.873), the two datasets where domain-aligned text supervision appears to
help most under scarce labels. H-Optimus-0 leads PCam (0.982, tied at three
decimals with UNI and GigaPath) and SPIDER-Breast (0.991), and UNI has the top CRC score (0.999,
tied at three decimals with H-Optimus-0). The largest discord with the
100\% regime appears on MHIST, where PLIP attains the highest AUROC (0.704)
despite ranking seventh of eight at full labels, and where MIDNIGHT falls to
0.499 (chance level). Every model collapses to a single predicted class on
MHIST at this fraction (macro-F1 0.387 for all eight), so these AUROC values
order the scores of a probe that makes one prediction. The cross-fraction
inversion in MHIST and BRACS rankings underlies the regime-aware FM selection
argument made in Section~\ref{sec:experiments}.

\begin{table}[!htbp]
\centering
\small
\caption{ECE (raw, before temperature scaling) at 1\% label fraction (mean $\pm$ std over 3 seeds). Lower is better. Best per dataset in bold.}
\label{tab:ece_raw_1}
\begin{tabular}{lcccccc}
\toprule
\textbf{Model} & \textbf{PCam} & \textbf{BRACS} & \textbf{CRC} & \textbf{MHIST} & \textbf{LungHist700} & \textbf{SPIDER-Breast} \\
\midrule
CLIP & 0.082\,$\pm$\,0.002 & \textbf{0.048\,$\pm$\,0.023} & 0.372\,$\pm$\,0.004 & 0.128\,$\pm$\,0.027 & \textbf{0.067\,$\pm$\,0.073} & \textbf{0.209\,$\pm$\,0.010} \\
PLIP & 0.072\,$\pm$\,0.004 & 0.069\,$\pm$\,0.042 & 0.287\,$\pm$\,0.009 & 0.146\,$\pm$\,0.025 & 0.307\,$\pm$\,0.051 & 0.364\,$\pm$\,0.008 \\
CONCH & \textbf{0.052\,$\pm$\,0.003} & 0.162\,$\pm$\,0.022 & 0.191\,$\pm$\,0.003 & 0.119\,$\pm$\,0.027 & 0.331\,$\pm$\,0.007 & 0.316\,$\pm$\,0.006 \\
VIRCHOW2 & 0.062\,$\pm$\,0.005 & 0.158\,$\pm$\,0.022 & 0.182\,$\pm$\,0.003 & 0.139\,$\pm$\,0.045 & 0.294\,$\pm$\,0.031 & 0.378\,$\pm$\,0.011 \\
UNI & 0.073\,$\pm$\,0.002 & 0.110\,$\pm$\,0.033 & 0.221\,$\pm$\,0.005 & 0.121\,$\pm$\,0.027 & 0.213\,$\pm$\,0.032 & 0.429\,$\pm$\,0.006 \\
GigaPath & 0.071\,$\pm$\,0.000 & 0.084\,$\pm$\,0.055 & 0.241\,$\pm$\,0.003 & 0.120\,$\pm$\,0.019 & 0.186\,$\pm$\,0.048 & 0.422\,$\pm$\,0.005 \\
H-Optimus-0 & 0.067\,$\pm$\,0.000 & 0.074\,$\pm$\,0.026 & 0.229\,$\pm$\,0.003 & 0.132\,$\pm$\,0.026 & 0.161\,$\pm$\,0.069 & 0.465\,$\pm$\,0.015 \\
MIDNIGHT & 0.057\,$\pm$\,0.004 & 0.106\,$\pm$\,0.009 & \textbf{0.182\,$\pm$\,0.001} & \textbf{0.091\,$\pm$\,0.004} & 0.173\,$\pm$\,0.113 & 0.310\,$\pm$\,0.010 \\
\bottomrule
\end{tabular}
\end{table}

Raw ECE at 1\% labels is largest on the multiclass datasets, and there it
reflects under-confidence of the regularized linear probe rather than
over-confidence: the validation-fitted temperature is below 1 in these cells
(Figure~\ref{fig:14_scaling_law}). On SPIDER-Breast (18 classes) the
pathology SSL models reach the highest raw ECE (H-Optimus-0 0.465, UNI 0.429,
GigaPath 0.422, VIRCHOW2 0.378), while CLIP, the non-pathology baseline, holds
the best score in the column (0.209). CLIP is also best on BRACS (0.048) and
LungHist700 (0.067), the same datasets on which it ranks last on AUROC, and
MIDNIGHT is best on MHIST (0.091) and CRC (0.182). A probe that commits weakly
to any class can be well calibrated in raw terms while discriminating poorly,
which reinforces the performance--reliability decoupling reported in
Figure~\ref{fig:7_decoupling}. These raw values use no validation labels;
after temperature scaling the ordering changes (Appendix tables and
Figure~\ref{fig:12_scaled_ece}).

\begin{table}[!htbp]
\centering
\small
\caption{ECE after post-hoc temperature scaling at 100\% label fraction. Lower is better.}
\label{tab:ece_scaled_100}
\begin{tabular}{lcccccc}
\toprule
\textbf{Model} & \textbf{PCam} & \textbf{BRACS} & \textbf{CRC} & \textbf{MHIST} & \textbf{LungHist700} & \textbf{SPIDER-Breast} \\
\midrule
CLIP & 0.040\,$\pm$\,0.000 & \textbf{0.018\,$\pm$\,0.000} & 0.004\,$\pm$\,0.000 & 0.107\,$\pm$\,0.000 & 0.176\,$\pm$\,0.000 & 0.029\,$\pm$\,0.000 \\
PLIP & 0.056\,$\pm$\,0.000 & 0.057\,$\pm$\,0.000 & 0.002\,$\pm$\,0.000 & 0.085\,$\pm$\,0.000 & 0.107\,$\pm$\,0.000 & 0.036\,$\pm$\,0.000 \\
CONCH & 0.037\,$\pm$\,0.000 & 0.025\,$\pm$\,0.000 & 0.002\,$\pm$\,0.000 & 0.085\,$\pm$\,0.000 & 0.104\,$\pm$\,0.000 & \textbf{0.025\,$\pm$\,0.000} \\
VIRCHOW2 & 0.035\,$\pm$\,0.000 & 0.050\,$\pm$\,0.000 & 0.002\,$\pm$\,0.000 & 0.062\,$\pm$\,0.000 & 0.102\,$\pm$\,0.000 & 0.027\,$\pm$\,0.000 \\
UNI & 0.041\,$\pm$\,0.000 & 0.081\,$\pm$\,0.000 & \textbf{0.001\,$\pm$\,0.000} & 0.061\,$\pm$\,0.000 & 0.122\,$\pm$\,0.000 & 0.032\,$\pm$\,0.000 \\
GigaPath & 0.039\,$\pm$\,0.000 & 0.062\,$\pm$\,0.000 & 0.002\,$\pm$\,0.000 & \textbf{0.054\,$\pm$\,0.000} & \textbf{0.078\,$\pm$\,0.000} & 0.036\,$\pm$\,0.000 \\
H-Optimus-0 & \textbf{0.030\,$\pm$\,0.000} & 0.050\,$\pm$\,0.000 & 0.001\,$\pm$\,0.000 & 0.057\,$\pm$\,0.000 & 0.088\,$\pm$\,0.000 & 0.028\,$\pm$\,0.000 \\
MIDNIGHT & 0.051\,$\pm$\,0.000 & 0.073\,$\pm$\,0.000 & 0.003\,$\pm$\,0.000 & 0.093\,$\pm$\,0.000 & 0.105\,$\pm$\,0.000 & 0.037\,$\pm$\,0.000 \\
\bottomrule
\end{tabular}
\end{table}

After post-hoc temperature scaling, in-distribution ECE drops below 0.10
for every (model, dataset) cell except for LungHist700 (the smallest
dataset, 691 tiles), where four of eight models still exceed 0.10. CRC is
trivially calibrated post-scaling, with all models reaching ECE
$\leq$\,0.004. The largest residuals after scaling concentrate on
LungHist700 (CLIP 0.176, UNI 0.122) and MHIST (CLIP 0.107), both
small-sample multiclass tasks where the validation split available for
fitting $T$ is thin. Combined with the raw values in
Table~\ref{tab:ece_raw_1}, this confirms the \emph{recoverability} axis
introduced in Section~\ref{sec:protocols}: a single scalar parameter
suffices for in-distribution calibration on five of six datasets, but
fails to repair OOD ECE (Table~\ref{tab:ood_ece_100}).

\begin{table}[!htbp]
\centering
\small
\caption{Brier score at 100\% label fraction. Lower is better.}
\label{tab:brier_100}
\begin{tabular}{lcccccc}
\toprule
\textbf{Model} & \textbf{PCam} & \textbf{BRACS} & \textbf{CRC} & \textbf{MHIST} & \textbf{LungHist700} & \textbf{SPIDER-Breast} \\
\midrule
CLIP & 0.118\,$\pm$\,0.000 & 0.095\,$\pm$\,0.000 & 0.010\,$\pm$\,0.000 & 0.168\,$\pm$\,0.000 & 0.110\,$\pm$\,0.000 & 0.024\,$\pm$\,0.000 \\
PLIP & 0.092\,$\pm$\,0.000 & 0.086\,$\pm$\,0.000 & 0.004\,$\pm$\,0.000 & 0.171 & 0.077 & 0.017\,$\pm$\,0.000 \\
CONCH & 0.074\,$\pm$\,0.000 & \textbf{0.072\,$\pm$\,0.000} & 0.003\,$\pm$\,0.000 & 0.161\,$\pm$\,0.000 & \textbf{0.052} & 0.014\,$\pm$\,0.000 \\
VIRCHOW2 & \textbf{0.047\,$\pm$\,0.000} & 0.072\,$\pm$\,0.000 & 0.001\,$\pm$\,0.000 & \textbf{0.130\,$\pm$\,0.000} & 0.055 & 0.008\,$\pm$\,0.000 \\
UNI & 0.053\,$\pm$\,0.000 & 0.074\,$\pm$\,0.000 & 0.001\,$\pm$\,0.000 & 0.142\,$\pm$\,0.000 & 0.062 & 0.008\,$\pm$\,0.000 \\
GigaPath & 0.053\,$\pm$\,0.000 & 0.073\,$\pm$\,0.000 & 0.001\,$\pm$\,0.000 & 0.141\,$\pm$\,0.000 & 0.064 & 0.008\,$\pm$\,0.000 \\
H-Optimus-0 & 0.048\,$\pm$\,0.000 & 0.074\,$\pm$\,0.000 & \textbf{0.001\,$\pm$\,0.000} & 0.143\,$\pm$\,0.000 & 0.067\,$\pm$\,0.000 & \textbf{0.007\,$\pm$\,0.000} \\
MIDNIGHT & 0.074\,$\pm$\,0.000 & 0.076\,$\pm$\,0.000 & 0.002\,$\pm$\,0.000 & 0.201\,$\pm$\,0.000 & 0.064 & 0.012\,$\pm$\,0.000 \\
\bottomrule
\end{tabular}
\end{table}

Brier score, a strictly proper scoring rule that combines discrimination
and calibration into one number, ranks the pathology vision SSL family
ahead of the vision-language models on most datasets, in line with the
AUROC patterns of Table~\ref{tab:auroc_100}. VIRCHOW2 leads on PCam
(0.047) and MHIST (0.130), CONCH on BRACS (0.072) and LungHist700 (0.052),
and H-Optimus-0 on CRC (0.001) and SPIDER-Breast (0.007). MIDNIGHT
records the highest Brier on MHIST (0.201), consistent with its weak
AUROC on the same dataset and confirming that the score reflects both
axes simultaneously. The very low Brier values on CRC (all models
$\leq$\,0.010) reflect the saturation already visible in AUROC and macro
F1, where the seven-class colorectal task admits near-perfect linear
separation in every embedding space considered.

\begin{table}[!htbp]
\centering
\small
\caption{Fitted temperature $T$ at 100\% labels, on the probe's own logit scale (binary values are the fitted two-column temperature divided by two). $T < 1$ indicates an under-confident probe, $T > 1$ an over-confident one; bold marks the value closest to 1 per dataset. Seeds agree exactly at 100\% labels.}
\label{tab:temperature_100}
\begin{tabular}{lcccccc}
\toprule
\textbf{Model} & \textbf{PCam} & \textbf{BRACS} & \textbf{CRC} & \textbf{MHIST} & \textbf{LungHist700} & \textbf{SPIDER-Breast} \\
\midrule
CLIP & \textbf{1.04} & 0.74 & 0.56 & 0.75 & 0.36 & 0.61 \\
PLIP & 1.07 & 0.78 & 0.65 & 0.72 & 0.41 & 0.64 \\
CONCH & 1.04 & 0.97 & \textbf{0.78} & \textbf{0.88} & 0.48 & \textbf{0.71} \\
VIRCHOW2 & 1.24 & 0.88 & 0.70 & 0.61 & 0.46 & 0.66 \\
UNI & 1.33 & 0.81 & 0.66 & 0.68 & 0.42 & 0.63 \\
GigaPath & 1.40 & 0.78 & 0.65 & 0.68 & 0.39 & 0.62 \\
H-Optimus-0 & 1.37 & 0.88 & 0.62 & 0.68 & 0.37 & 0.66 \\
MIDNIGHT & 1.19 & \textbf{0.98} & 0.72 & 0.86 & \textbf{0.51} & 0.60 \\
\bottomrule
\end{tabular}
\end{table}

At full supervision the direction of miscalibration is set by the dataset
rather than by the model. On the four multiclass datasets every FM is
under-confident ($T < 1$; as low as 0.36 on LungHist700). On PCam every FM is
mildly over-confident ($T$ from 1.04 to 1.40), whereas on MHIST all FMs are
under-confident ($T$ from 0.61 to 0.88). Across all 288 cells the probe is
under-confident in 252, which is consistent with a strongly regularized linear
probe on L2-normalized features producing compressed logits.

\begin{table}[!htbp]
\centering
\small
\caption{Macro-F1 at 100\% label fraction (seeds agree exactly). Best per dataset in bold.}
\label{tab:f1_100}
\begin{tabular}{lcccccc}
\toprule
\textbf{Model} & \textbf{PCam} & \textbf{BRACS} & \textbf{CRC} & \textbf{MHIST} & \textbf{LungHist700} & \textbf{SPIDER-Breast} \\
\midrule
CLIP & 0.831 & 0.433 & 0.950 & 0.654 & 0.216 & 0.647 \\
PLIP & 0.872 & 0.498 & 0.976 & 0.642 & 0.597 & 0.738 \\
CONCH & 0.898 & 0.589 & 0.983 & 0.713 & 0.701 & 0.783 \\
VIRCHOW2 & 0.937 & 0.596 & 0.995 & \textbf{0.815} & \textbf{0.760} & 0.873 \\
UNI & 0.935 & 0.596 & 0.995 & 0.773 & 0.753 & 0.875 \\
GigaPath & 0.934 & \textbf{0.605} & 0.994 & 0.785 & 0.725 & 0.866 \\
H-Optimus-0 & \textbf{0.939} & 0.585 & \textbf{0.996} & 0.773 & 0.737 & \textbf{0.888} \\
MIDNIGHT & 0.906 & 0.561 & 0.988 & 0.540 & 0.681 & 0.813 \\
\bottomrule
\end{tabular}
\end{table}

Macro-F1 at full label availability separates the eight FMs more sharply
than AUROC because it depends on the decision threshold and on minority-class
recall. H-Optimus-0 leads on PCam (0.939), CRC (0.996) and SPIDER-Breast (0.888), VIRCHOW2 on MHIST (0.815) and LungHist700 (0.760), and GigaPath on BRACS (0.605). MIDNIGHT records the lowest macro-F1 on MHIST (0.540), mirroring its low AUROC there. The CLIP gap relative to the pathology-pretrained FMs is much larger in macro-F1 (BRACS 0.433, LungHist700 0.216) than in AUROC, which motivates the breaking-point analysis in Figure~\ref{fig:4_breaking_point}.

\section{OOD transfer detailed results}
\label{app:ood_tables}
Per-(model, transfer pair) numerical OOD results corresponding to Figures~\ref{fig:10_ood_ece}, \ref{fig:8_ood_auroc_curves} and \ref{fig:9_ood_auroc_drop}.
\begin{table}[!htbp]
\centering
\small
\caption{OOD AUROC on target dataset, trained at 100\% source labels. Best per transfer pair in bold.}
\label{tab:ood_auroc_100}
\begin{tabular}{lcccc}
\toprule
\textbf{Model} & \textbf{PCam$\to$MHIST} & \textbf{MHIST$\to$PCam} & \textbf{CRC$\to$BRACS} & \textbf{BRACS$\to$CRC} \\
\midrule
CLIP & 0.484 & 0.628 & \textbf{0.661} & 0.527 \\
PLIP & 0.548 & 0.575 & 0.613 & 0.434 \\
CONCH & 0.633\,$\pm$\,0.000 & \textbf{0.696} & 0.428 & 0.240 \\
VIRCHOW2 & 0.585 & 0.685\,$\pm$\,0.000 & 0.635 & \textbf{0.746} \\
UNI & 0.442 & 0.589 & 0.523 & 0.401 \\
GigaPath & 0.588 & 0.487 & 0.602 & 0.602 \\
H-Optimus-0 & \textbf{0.645} & 0.561\,$\pm$\,0.000 & 0.574 & 0.501 \\
MIDNIGHT & 0.445 & 0.430\,$\pm$\,0.000 & 0.454 & 0.622 \\
\bottomrule
\end{tabular}
\end{table}

OOD AUROC values reveal that several pathology FMs lose discrimination
under cross-domain shift even at full source-label availability. On
PCam$\to$MHIST H-Optimus-0 leads (0.645) but CLIP, MIDNIGHT, and UNI
remain near or below 0.5; on MHIST$\to$PCam CONCH leads (0.696) and
MIDNIGHT collapses to 0.430. The most striking entry is CONCH on
BRACS$\to$CRC, with OOD AUROC = 0.240, well below the random-prediction
threshold of 0.5, indicating systematically inverted predictions on the
target taxonomy. Comparison with Table~\ref{tab:auroc_100} makes the
in-distribution to OOD gap explicit: a model with ID AUROC of 0.999 on
CRC drops to 0.240 in the BRACS$\to$CRC transfer, a magnitude of
degradation that is invisible in any single-dataset benchmark.

\begin{table}[!htbp]
\centering
\small
\caption{OOD ECE (raw, 15 equal-width bins) on the target dataset at 100\% source labels (seeds agree exactly). Lower is better.}
\label{tab:ood_ece_100}
\begin{tabular}{lcccc}
\toprule
\textbf{Model} & \textbf{PCam$\to$MHIST} & \textbf{MHIST$\to$PCam} & \textbf{CRC$\to$BRACS} & \textbf{BRACS$\to$CRC} \\
\midrule
CLIP & 0.251 & 0.312 & 0.204 & 0.300 \\
PLIP & 0.261 & \textbf{0.207} & \textbf{0.179} & 0.261 \\
CONCH & 0.151 & 0.455 & 0.338 & 0.261 \\
VIRCHOW2 & \textbf{0.123} & 0.453 & 0.268 & \textbf{0.138} \\
UNI & 0.285 & 0.439 & 0.272 & 0.188 \\
GigaPath & 0.288 & 0.462 & 0.232 & 0.278 \\
H-Optimus-0 & 0.316 & 0.488 & 0.275 & 0.220 \\
MIDNIGHT & 0.362 & 0.304 & 0.268 & 0.218 \\
\bottomrule
\end{tabular}
\end{table}

OOD ECE on the target is several times larger than the corresponding
in-distribution ECE in Table~\ref{tab:ece_scaled_100}. The MHIST$\to$PCam
column is the most extreme: H-Optimus-0 0.488, GigaPath 0.462, CONCH 0.455, VIRCHOW2 0.453, on a target where the same models have
in-distribution ECE below 0.06 after temperature scaling. PLIP has the lowest OOD ECE on MHIST$\to$PCam (0.207) and CRC$\to$BRACS (0.179), inverting the in-distribution ranking on those pairs, while VIRCHOW2 is best on PCam$\to$MHIST (0.123) and BRACS$\to$CRC (0.138). These are label-definition pairs on which every model transfers near chance (Table~\ref{tab:ood_auroc_100}), so the values describe calibration of failing transfer rather than a ranking of FMs.

\end{document}